\def\year{2022}\relax
\documentclass[letterpaper]{article} 
\usepackage{aaai22}  
\usepackage{times}  
\usepackage{helvet}  
\usepackage{courier}  
\usepackage[hyphens]{url}  
\usepackage{graphicx} 
\usepackage{natbib}  
\usepackage{caption} 
\DeclareCaptionStyle{ruled}{labelfont=normalfont,labelsep=colon,strut=off} 
\usepackage{algorithm}
\usepackage{algorithmic}

\usepackage[colorlinks, 
            linkcolor=blue, 
            anchorcolor=blue,
            citecolor=blue]{hyperref}
\usepackage{hhline}
\usepackage{array} 
\usepackage[encapsulated]{CJK}
\usepackage{booktabs}
\usepackage{tabularx}
\usepackage{mathrsfs}
\usepackage{amsmath}
\usepackage{dsfont}
\usepackage[table]{xcolor}
\usepackage{float}
\usepackage{placeins}
\usepackage[titletoc]{appendix}
\usepackage{bbold}
\usepackage{multirow}
\usepackage{cleveref}  

\usepackage{mdframed}
\newmdenv[
  skipabove=\topsep,
  skipbelow=\topsep
]{siderules}

\usepackage{newfloat}
\usepackage{listings}
\floatstyle{ruled}
\newfloat{listing}{tb}{lst}{}
\floatname{listing}{Listing}
\title{\textsc{AspirinSum}: an \textsc{Asp}ect-based ut\textsc{i}lity-p\textsc{r}eserved de-\textsc{i}de\textsc{n}tification \textsc{Sum}marization framework}
\author{
    Ya-Lun Li\textsuperscript{\rm 1}
}
\affiliations{
    \textsuperscript{1}Institute of Information Systems and Applications\\
    National Tsing Hua University\\

    Hsinchu, Taiwan, R.O.C.\\
    \textsuperscript{1}allen7575@gmail.com
}

\begin{document}
\begin{CJK}{UTF8}{bsmi}
\definecolor{important_green}{RGB}{79, 173, 91}
\definecolor{background_blue}{RGB}{47, 110, 186}
\definecolor{unique_orange}{RGB}{222, 131, 68}

\pagestyle{plain} 

\pagenumbering{arabic} 

\maketitle

\begin{abstract}

Due to the rapid advancement of Large Language Model (LLM), the whole community eagerly consumes any available text data in order to train the LLM. Currently, large portion of the available text data are collected from internet, which has been thought as a cheap source of the training data. However, when people try to extend the LLM's capability to the personal related domain, such as healthcare or education, the lack of public dataset in these domains make the adaption of the LLM in such domains much slower. The reason of lacking public available dataset in such domains is because they usually contain personal sensitive information. In order to comply with privacy law, the data in such domains need to be de-identified before any kind of dissemination. It had been much research tried to address this problem for the image or tabular data. However, there was limited research on the efficient and general de-identification method for text data. Most of the method based on human annotation or predefined category list. It usually can not be easily adapted to specific domains. The goal of this proposal is to develop a text de-identification framework, which can be easily adapted to the specific domain, leverage the existing expert knowledge without further human annotation. We propose an aspect-based utility-preserved de-identification summarization framework, \textsc{AspirinSum}, by learning to align expert's aspect from existing comment data, it can efficiently summarize the personal sensitive document by extracting personal sensitive aspect related sub-sentence and de-identify it by substituting it with similar aspect sub-sentence. We envision that the de-identified text can then be used in data publishing, eventually publishing our de-identified dataset for downstream task use.

\end{abstract}


\newpage
\tableofcontents
\newpage

\section{Introduction}
\label{sec:introduction}

Recently, due to the advancement of Large Language Model (LLM), it can be utilized in various domains, which like healthcare~\cite{Fox2023, Leonard2023}, finance~\cite{Estrada2023, Ferreira2023, taver2023chatgpt}, legal~\cite{Black2023legalassistants, Braff2023ChatGPT}, education~\cite{JAVAID2023100115}, personal counseling~\cite{Germain2023, kimmel2023chatgpt}, ...etc. However, training a domain specific LLM needs large amount of domain specific data, which usually can not available due to the lack of public released dataset. The reason of lacking domain specific public dataset is mainly because the domain specific data usually contains personal sensitive information which can be utilized to identify individuals. Releasing such data without proper de-identification process may leak personal information to the public, hence violate privacy protection law. 

Conversely, LLMs has been proved that it is prone to the training data extraction attack even when it has been well aligned~\cite{nasr2023scalable, 274574, lehman2021does}. Imagine that a hospital or a school use private data to fine-tune a self-hosted open source LLM and use it to serve patients~\cite{mukherjee2023feasibility} or students. The attacker may find a special prompt to ask the LLM to print out the training data, which is expected to be private and should not be exposed, causing huge privacy concern. Therefore, the most secure way seems to de-identify the training data before it is used to train the model~\cite{brown2022does}.

Current textual data de-identification method heavily rely on sequence labeling based model, it focuses on detecting sensitive text and redacting it. While the state-of-the art model~\cite{anjum2021identification} already achieve very high recall rate (about \~98.91\%), it still suffers from the remaining 1\% missing rate. Due to the huge amount of the data, even a small portion of private text leakage can lead to the re-identification of the patients, make 99\% recall rate ineffective. There is a study suggested that the permissible missing rate should be lower than 1\%~\cite{yogarajan2020review}, which is usually very hard to achieve. The other drawback of sequence labeling based model is that it is struggled to deal with diverse types of indirect identifiers.

Traditionally, in order to detect Personally Identifiable Information (PII), we need first define a list of sensitive identifier types that needs to be removed, then asking human annotators to follow this type definitions to produce label data. For example, Table~\ref{tab:pii_types} shows that HIPAA~\cite{annas2003hipaa} defined PII types. However, there are many other indirect identifiers can be used to identify individuals without any proper definition. For example, in the educational field, a student's special experiences and unique achievements can be used to identify that student, but the lacking of a unified definition of such indirect identifiers makes it hard to ask humans to label it. Even worse, the creation of a comprehensive list of indirect identifier types is also hard due to its ambiguity, variety and diversity. Furthermore, studies~\citep{staab2023beyond, patsakis2023man} had shown that LLMs have the ability to re-identify or infer individual's PII solely based on nuance textual clues even remove the conventional defined PII identifiers.

\begin{figure}[htbp]
\centerline{\includegraphics[width=0.5\textwidth]{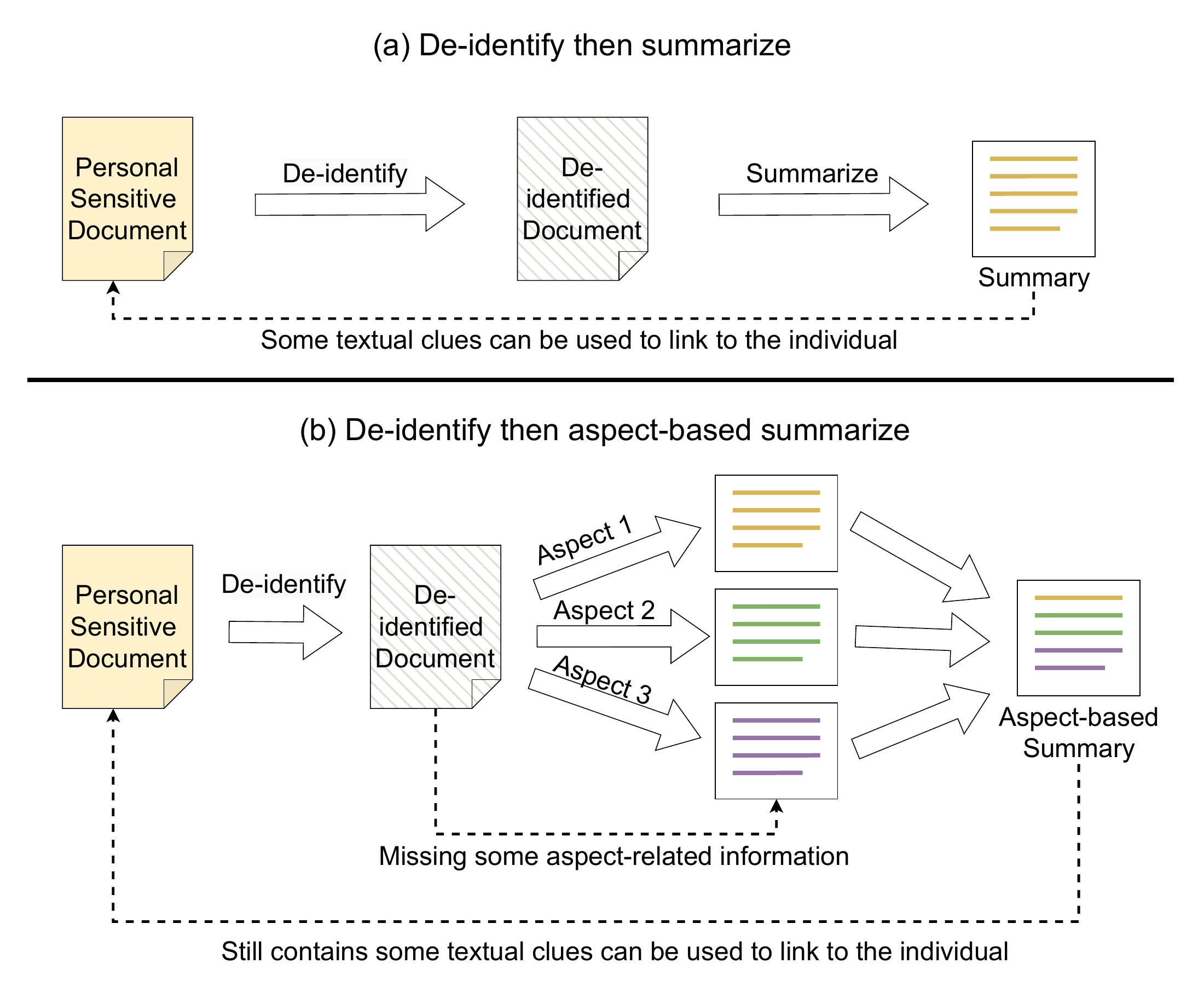}}
\caption{Conventional methods}
\label{fig:conventional_methods}
\end{figure}

On the other hand, a small but representative subset of the training data can be used to fine-tuning LLMs more efficiently~\cite{das2023deft}. In order to create a small subset for fine-tuning purpose, we might want to create a summarized version of the original dataset. In our case, we want to create a summary dataset of the original documents that can be used in fine-tuning other models without leaking sensitive information. In order to do that, a naive method is to run conventional PII remover on the original document first, then send it to the general summarization model, as shown in Figure~\ref{fig:conventional_methods}(a). However, the generated summary might be too general to be used in the downstream task and still contain several textual clues that can be used to link to the original document owner. 

\begin{figure}[htbp]
\centerline{\includegraphics[width=0.5\textwidth]{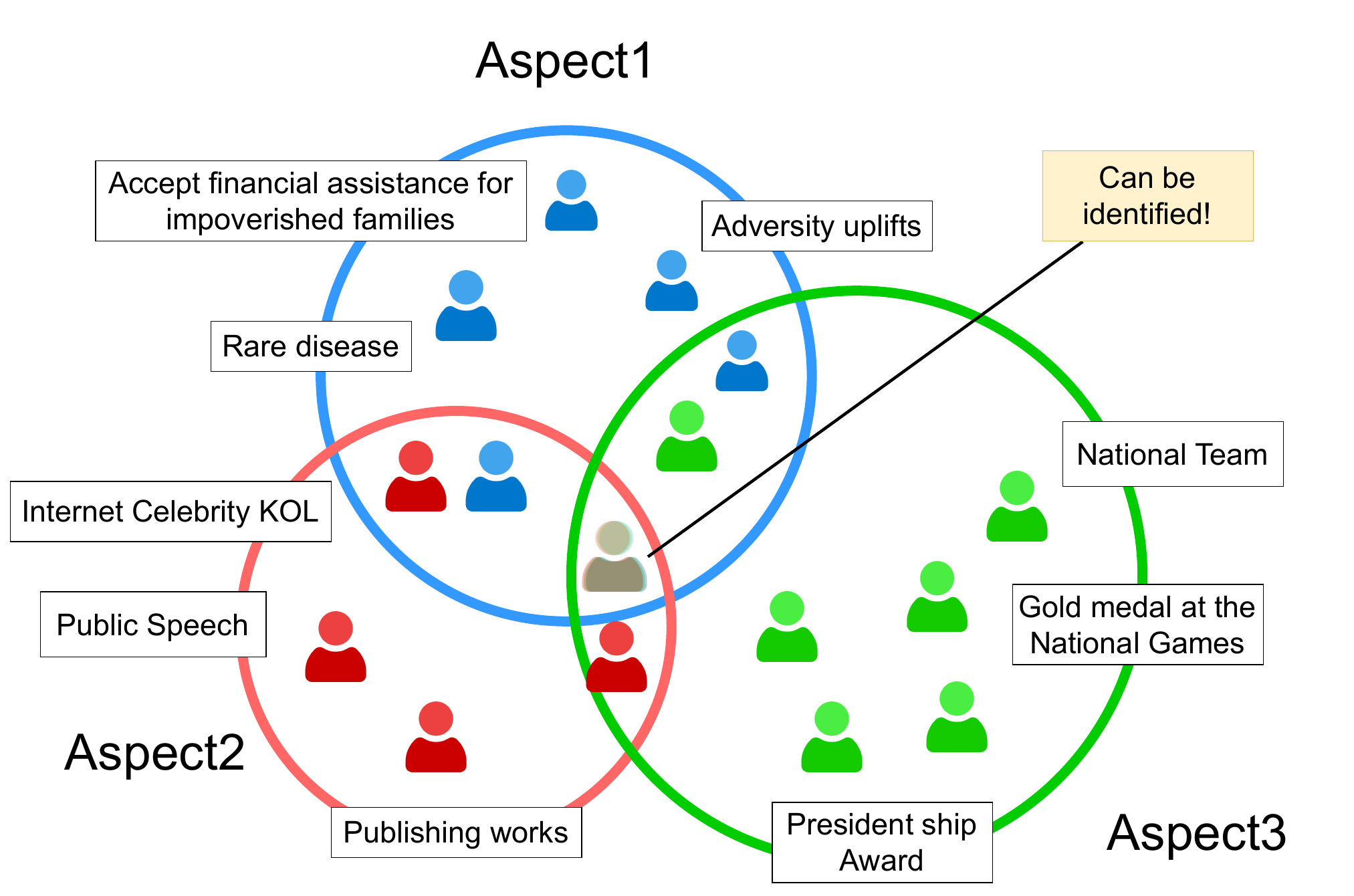}}
\caption{Personal Sensitive Aspect}
\label{fig:personal_sensitive_aspect}
\end{figure}

To avoid the generated summaries too general, there were several multi-perspective summarization methods had been proposed. The multi-perspective summarization method tries to extract the most salient aspects across the document, generate a summary for each aspect, and aggregate them into a final summary, as shown in Figure~\ref{fig:conventional_methods}(b). When the multi-perspective summarization method is applied to the PII removed document, however, some useful information might be removed, making the generated multi-perspective summary less useful. This occurs especially when the salient aspects are also the personal sensitive information, we call these aspects \textbf{Personal Sensitive Aspect (PSA)}, as shown in Figure~\ref{fig:personal_sensitive_aspect}.

In this study, we aim to propose an aspect-based machine summarization framework that can generate de-identified summaries for personal sensitive documents. The goal is to remove the linkage to the document owner from the generated summary while maintaining its utility for downstream tasks, ultimately creating a publish-ready summary dataset for the high-school student's college applications.

Our envisioned contributions are four folds:

\begin{enumerate}
\item We proposed a reference text guided, aspect-based utility-preserved de-identification summarization framework, called \textsc{AspirinSum}, which can easily be applied to other privacy-sensitive summarization tasks without losing useful information at aspect level.

\item We formulate the de-identification problem as an aspect discovery task, and demonstrate that it can be a more flexible method than the traditional NER-based method, without using a fixed set of pre-defined PII categories.

\item We thoroughly investigate the utility, fidelity and re-identifiability of the generated de-identify summary, by proposing several new scoring metrics. 

\item We plan to releasing the de-identified summary dataset called \textbf{High School Student's College Application Summary (HSSCAS)} dataset, which utilize our method to summarize student application document and de-identify sensitive information, while keep its admission score related aspect information for downstream task use. 
\end{enumerate}

\begin{table}[ht]
\centering
\begin{tabular}{ccm{5cm}}
\hhline{===}
\textbf{No.} & \textbf{PHI Type} & \multicolumn{1}{c}{\textbf{Description}} \\
\hhline{===}
1 & Names & First, last, hospital names \\
2 & Location & Any geographic divisions smaller than a state \\
3 & Dates & Birth date, admission or discharge date, etc. \\
4 & Contact & Home, office, or cell phone numbers \\
5 & Vehicle & Vehicle serial or license plate numbers \\
6 & Fax & Fax information \\
7 & Device & Device identifiers and serial numbers \\
8 & Email & Any electronic mail addresses \\
9 & URLs & Web Universal Resource Locators \\
10 & SSNs/SINs & Social security or insurance number \\
11 & MRNs & Medical record numbers \\
12 & IP & Internet Protocol address \\
13 & Biometric & All finger or voice-prints \\
14 & Insurance & Health plan beneficiary numbers \\
15 & Photo & Full-face (or similar) photographic images \\
16 & Accounts & Bank accounts, social media profile \\
17 & Certificate & License or certificate number \\
18 & ID & Any unique identifying numbers \\
\hhline{===}
\end{tabular}
\caption{HIPAA Safe Harbor Method Defined 18 PII Types ~\citep{garfinkel2015identification}}
\label{tab:pii_types}
\end{table}



\subsection{Problem Overview}
\label{subsec:problem_overview}

Researchers have collected a vast amount of text data from various sources, including public ones like Wikipedia and private ones like proprietary books, to train a general-purpose Large Language Model (LLM). However, such LLMs are not yet well-suited for domain-specific tasks, particularly in sensitive areas like healthcare, education, or personal counseling. These domains typically involve Personal Identifiable Information (PII), which must be removed to comply with privacy laws. However, removing PII often requires costly human labeling and a clear list of categories to be removed, which may not always be available for a given domain. Consequently, there are few publicly available domain-specific datasets suitable for LLM training, limiting its usage in domain-specific tasks such as expert-aware aspect-based summarization, answering domain-specific questions without compromising sensitive information, and generating utility-aware outputs for downstream tasks. Addressing these challenges requires a systematic framework for efficiently publishing domain-specific datasets that are expert-aware, de-identified, and utility-preserved. Key challenges include (a) dataset acquisition and ethical concerns, (b) expert-aware domain-specific summarization, (c) removal of individual linkages, and (d) dataset publication and downstream task utility. Further details will be provided in the following paragraphs.

\subsubsection{Dataset Acquisition and Ethical Concerns}

In areas like healthcare and education, institutions often gather large amounts of data, such as clinical notes or college applications, but accessing this data for research can be challenging due to privacy concerns. Researchers typically need to undergo a research ethics review, such as with an Institutional Review Board (IRB), to ensure ethical use and compliance with privacy laws. In this study, we seek access to college application data from high school students who applied to the Special Talent Admission Program. These data include students' self-statements and committee members' comments, which could potentially identify individuals when combined with other personal information from sources like social media or newspapers. To make this dataset available for research while protecting privacy, we propose an aspect-based de-identification summarization framework. Details of the data processing mechanism are provided in later sections.

\subsubsection{Expert-aware Domain-specific Summarization}

While recent advances in general-purpose LLMs excel at general text summarization, they may fall short in specific domains by overlooking key aspects known to domain experts. For instance, in the field of education, admission screening entails considering various perspectives from experts in diverse fields to evaluate applicants comprehensively. LLMs, as general-purpose agents, lack detailed knowledge of each field unless fine-tuned with domain expert opinion data. However, gathering such data is typically costly and time-consuming. Conversely, collecting domain expert notes, often containing crucial points in abbreviated form or brief reasoning, from real-world records like clinical notes or admission committee comments is more feasible. This proposal aims to utilize such domain expert notes to generate expert-aware, aspect-based summaries from the given domain data requiring summarization, facilitating downstream task utilization.

\subsubsection{Removal of Individual Linkages}

Many current summarization frameworks overlook the need to de-identify documents while preserving their utility. They typically prioritize faithful summarizing the content. However, when a document contains sensitive personal information, the resulting summary may unintended expose these details, raising privacy concerns. Conversely, humans excel at summarizing personal information in a non-identifiable manner, often by imagining or fabricating scenarios that capture essential aspects, akin to alluding or insinuating without directly mentioning personal details.

Unlike HIPAA, which provides a widely used list of PII categories (Table~\ref{tab:pii_types}) in the healthcare domain, there is no well-defined public list of sensitive categories in other domains such as education. Because the content of personal sensitive information can vary depending on the context, it is challenging to create a predefined list of categories that covers all possible sensitive aspects across different contexts.

Instead of creating a list of PII categories, labeling PII tokens, and deleting them from the document, this proposal aims to utilize all available reference data, such as doctors' clinical notes or committee members' comments, to identify Personal Sensitive Aspects (PSA) (see Figure~\ref{fig:personal_sensitive_aspect}). This is because domain experts often introduce terms in these reference notes that are not only salient to the person they evaluate but also capable of identifying the individual, such as events, achievements, awards, medical treatments, etc. By identifying these personal sensitive aspect terms, we can replace them with other similar utility terms that are not linked to the individual, thus breaking connections to specific individuals.

\subsubsection{Dataset Publication and Downstream Task Utility}

When sharing sensitive, domain-specific data with researchers and practitioners for tasks like demographic analysis, machine classification, or fine-tuning LLMs, it's often more effective to offer a summarized version rather than the original. This is because summaries are more concise and contain denser information. To create such summaries, it's crucial to retain the most representative information from the original data for downstream tasks. 

However, in fields like healthcare or education, the representative information itself is often sensitive and can be used to identify individuals. Simply removing this sensitive information, as conventional anonymization tools do, might reduce the utility of the resulting summary since it also removes representative information necessary for downstream tasks. To release the dataset, it requires achieving both data anonymity and utility simultaneously. But these two objectives often conflict, making dataset release more challenging.

Instead of aiming to faithfully summarize individuals, this proposal suggests generating synthesized summaries by randomly selecting similar aspect sub-sentences from groups of similar peoples to replace the original sub-sentences. This approach removes links to specific individuals while preserving useful information for publication without sacrificing utility.

\subsection{Research Questions}
\label{subsec:research_questions}

In the previous section, we identified and discussed various challenges that haven't been addressed in the existing literature. We then summarized these challenges in several research questions (RQs) as follows. These RQs represent the primary objectives of our proposal to publish a dataset that is expert-aware, domain-specific, de-identified, and preserves utility.

\textbf{RQ1:} Can we get high-quality labeled data for domain-specific summarization which is aware of human expert's aspects? If not, can we use noisy reference data, such as expert's comments or notes, as a proxy of the label data?

\textbf{RQ2:} How can we train a model to align with experts' opinions from noisy reference data, and can be used to infer on unseen data?

\textbf{RQ3:} Can we generate summaries that are as effective for the downstream task as possible while preventing any direct or indirect personal identifiable information from leaking?

\textbf{RQ4:} How can we remove linkages between individuals and generated summaries without relying on any predefined domain-specific category lists (such as HIPAA)?

\section{Related Work}
\label{sec:related_Work}

In this section, we examine different aspect-based summarization methods employed in previous studies for extracting key aspects from reference data such as product reviews. Additionally, we discuss various privacy-preserving methods proposed to prevent PII leakage in textual data. The review covers the details of multi-perspective summarization, unsupervised aspect extraction, and techniques for text de-identification and anonymization.

\subsection{Multi-Perspective Summarization}
\label{subsec:multi_perspective_summarization}

In the traditional generic text summarization method, no matter \textit{extractive}~\cite{luhn1958automatic, gong2001generic, steinberger2004using, mihalcea2004textrank, erkan2004lexrank, bougouin2013topicrank, florescu2017positionrank, chengzhang2018chinese, haider2020automatic, abdulateef2020multidocument, giarelis2023abstractive} or \textit{abstractive}~\cite{ganesan2010opinosis, genest2012fully, khan2018abstractive, rekabdar2019generative, yang2020hierarchical, raffel2020exploring, lewis2019bart, zhang2020pegasus, brown2020language}, they are focusing on single perspective summarization, which means, for each document to be summarized, it usually needs a single version of human written gold standard summary as a reference summary to evaluate how good the generated summaries are. But in real world cases, most of the time, each person can have their own perspective to see the document, each cover several different aspects, hence a single version of gold standard summary may not reflect the real world cases.

The goal of \textit{multi-perspective summarization} is trying to generate a summary which cover the aspects of a target item or document as more as possible. In order to model the possible aspects of interest from real world data, a natural source is the online product reviews, which usually contains various opinions coming from different person. Due to the diversity of opinions, it is very hard to ask human annotators to write a gold standard summary for each product or document. Also, in domains which like healthcare, finance, legal or education, we don't have any aspect labeled data, so the method to automatically discover aspects from opinions is very crucial. Therefore, the multi-perspective summarization usually formulated as an unsupervised task, which encompasses two sub-tasks: 1. \textit{aspect discovery} and 2. \textit{aspect-based summarization}. 

\subsubsection{Aspect discovery}
There have been proposed several different ways to solve the aspect discovery problem, including rule-based, supervised, unsupervised, and weakly supervised. \textit{Rule-based approaches} utilize a set of manually defined lexicon patterns to find aspects~\cite{qiu2011opinion, liu2016improving}, which usually needs to incorporate domain knowledge and human expertise. \textit{Supervised approaches} usually formulate aspect extraction as a sequence labeling problem, which can be solved by hidden Markov models (HMM)~\cite{jin2009opinionminer}, conditional random field (CRFs)~\cite{li2010structure, mitchell2013open, yang2012extracting} and recurrent neural network (RNN)~\cite{wang2016recursive, liu2015fine}. Although its better performance compares to the rule-based approaches, it requires a large amount of labeled data for training. Early attempts of \textit{Unsupervised approaches}, which like Latent Dirichlet Allocation (LDA)-based topic modeling~\cite{brody2010unsupervised, zhao2010jointly, chen2014aspect, garcia2018w2vlda, shi2018short}, Restricted Boltzmann Machine (RBM)~\cite{wang2015sentiment}, do not need labeled data, but suffer from its lower performance compare to supervised methods.

Recently, deep learning based models have shown strong performance in extracting coherent aspects.~\citet{he2017unsupervised} first proposed an unsupervised autoencoder aspect extraction framework, called ABAE, which can automatically discover aspects without any supervision. Although it's unsupervised, it still needs to manually assign discovered aspects to the top ranked representative words, and needs to set larger number of topics to be discovered ($\sim$15) compare to the actual aspect found in the data ($\sim$5). Base on ABAE,~\citet{angelidis2018summarizing} proposed MATE with several improvements. First, they introduce \textit{Multi-Seed Aspect Extraction}, by annotate a small set of seed words ($\sim$30) for each aspect, it can capture more meaningful aspects. Second, they incorporate sentiment polarity classification task to ranking each opinion, this helps to choose salience opinions. Third, they split sentence into \textit{Elementary Discourse Unit} (EDU)~\cite{mann1988rhetorical}, which facilitates the performance of summarization~\cite{li2016role}. Instead of using human annotated aspect seed-words,~\citet{zhao2020weakly} proposed \textsc{AspMem}, they leverage products' feature description, which can be easily found on the product's webpage, to collect the seed-words. In order to extract seed-words for each product category, they apply TF-IDF to the single document of the concatenated product descriptions of the same category, and select top $K$ words from this document as the seed-words of the product category. By taking the average word embedding of seed-words as the aspect embedding, they can identify the aspect of each review segment by calculating their cosine similarity. However, unlike product review, which usually can obtain the product category, the data we consider may not contain knowledge about categories, making this method hard to apply to general cases.

\citet{shi2021simple} proposed a self-supervised contrastive learning framework. By modeling aspect as an aggregation of neighbor word embeddings, it can learn a reasonable aspect embedding with a novel attention module called Smooth Self-Attention (SSA). Contrary to the regular self-attention, which can only capture single keyword, SSA can capture phrase and multiple keywords in the text segments, hence learn robust aspect representations. They also utilize knowledge distillation technique to train a student classifier from their semi-manual aspect mapping function to avoid the noise introduced by data pre-processing, resulted better performance in segment aspect predictions.

\citet{angelidis2021extractive} proposed Quantized Transformer, which is inspired by Vector-Quantized Variational Autoencoder (VQ-VAE)~\cite{van2017neural}. It can automatically cluster sentences which have similar aspect with similar latent codes. To avoid a sentence contains multiple aspects cause it never being the nearest to a cluster center, they introduce a sentence sampling method, which first sample a cluster, then sample sentences around that cluster. They also provide a newly created dataset called \textsc{Space}, a large-scale opinion summarization benchmark for evaluation of unsupervised summarizers. Instead of using hard vector quantized latent code, \citet{chowdhury2022unsupervised} proposed SemAE, which use soft distribution with sparse induced regularization loss to obtain sparse aspect representation.~\citet{amplayo2021aspect} proposed \textsc{AceSum}, which utilize \textit{Multiple Instance Learning} (MIL) \cite{keeler1991self} along with multi-head attention as pooling method to obtain token-level, sentence-level and document-level aspect labels. By utilized this inferred aspect code, they can query with single or multiple aspect code to generate aspect-specific summaries.

\citet{fabbri2021multi} proposed a novel multi-perspective dataset creation method, by clustering relevant sentences, extracting cluster centroids, use these centroids as bullet point summaries, each bullet point represents a perspective. They use Reinforcement Learning (RL) with three reward functions to fine-tune the BART~\cite{lewis2019bart} model: ROUGE~\cite{lin2004rouge} for content coverage, sentence-level entailment~\cite{falke2019ranking} for faithfulness and semantic area ~\cite{yogatama2015extractive, jung2019earlier} for perspective coverage.

\citet{zhao2023cone} proposed \textsc{Cone}, an unsupervised aspect extraction method. They first give sentences aspect and sentiment pseudo labels by k-means clustering and a rule-based sentiment classifier, then do the contrastive learning (CL). In order to do the CL, they construct positive pair by backtranslation, negative pair by randomly sampling sentence with different pseudo aspect or sentiment from different documents, update the latent representation of aspect and sentiment seperatly. After CL, they do the k-means cluster again, to refine the pseudo label. By interative refine and updating the cluster, the extracted aspect and sentiment can be improved. However, the unsupervised clustering based method usually suffer from extract redundant aspect and mix of related aspects, due to the fact that the unsupervised learning can not produce disentangled representation in a human meaningful way without inductive bias~\cite{locatello2019challenging}.

\subsubsection{Aspect-based summarization} \citet{jiang2023large} proposed \textsc{SubSumm}, a large-scale and multi-perspective opinion summarization framework. They present a review sampling strategy which based on sentiment analysis and contrastive information valuation to select a small but valuable review subset. By using this small subset of reviews as input, they fine-tune a pre-trained language model BART~\cite{lewis2019bart} to produce the multi-perspective summary. However, it needs dataset to provide reference summary in different perspectives as training target, but in most cases, we don't have such reference summary, this limits the usage of this method.

\citet{suhara2020opiniondigest} proposed \textsc{OpinionDigest} framework, by utilizing pre-trained Aspect-Based Sentiment Analysis (ABSA) tagging
model~\cite{miao2020snippext} to extract opinion phrases, training the Transformer~\cite{vaswani2017attention} model to reconstruct review text from extracted opinion phrases.~\citet{zhang2023asu} proposed AsU-OSum, a graph-based aspect-augmented unsupervised opinion summarization framework. They also utilize a pre-trained ABSA tagging model~\cite{miao2020snippext} to extract aspects and opinions. By clustering similar aspects of opinions, they can construct a knowledge-graph connecting opinion nodes and review nodes, then utilize graph attention network (GAT)~\cite{velivckovic2017graph} to obtain node's representation to generate aspect-augmented summary. However, it still needs to use labeled reviews data to fine-tune the ABSA tagging model. In many cases, it is hard to ask human expert to label domain specific data in ABSA aware manner, makes these ABSA-based method inapplicable.

\citet{hsiao2022generate} proposed \textit{Pairwise Alignment Mechanism} to align facts and reasons which appearing in the legal document to extract multiple perspectives. They use Hierarchical Attention Network (HAN)~\cite{yang2016hierarchical} to obtain the attention score for each word, and combine with its POS tagging to filter out redundant words and extract explicit keywords. They obtain the \textit{Explicit Alignment Score} by calculating Jaccard similarity~\cite{levandowsky1971distance} of the extracted keyword set between facts and reasons. They also calculate cosine similarity between the embedding of facts and reasons to obtain the \textit{Implicit Alignment Score}. Combining these two alignment scores, they can extract perspectives present in facts and its corresponding reasons.

More recently, \citet{bhaskar-etal-2023-prompted} proposed a LLM-based opinions summarization method, they first prompted GPT-3.5 to describe the topic of each sentence of reviews in one word, utilize these topic words to group sentences by topic, and prompted GPT-3.5 to summarize what reviewer says per aspect. This method is zero-shot without any training. However, the performance of zero or few-shot prompt based method is heavily dependents on LLM's pre-train data. If the pre-train data doesn't include the target domain data, the inferred topic words may not align with the aspect of expert interest in the target domain. 

\subsection{Privacy-Preserving Methods for Text Data}
\label{subsec:privacy_preserving_methods_for_text_data}

The Personally Identifiable Information (PII)~\cite{garfinkel2015identification} is the most important concept in privacy regulation laws which like HIPAA~\cite{annas2003hipaa}, PIPEDA~\cite{nisker2006pipeda} and GDPR~\cite{http://data.europa.eu/eli/reg/2016/679/oj}. It defines the information that need to be removed from data before dissemination to avoid any linkage to the individuals 
and hence violate privacy protection laws. There are two types of PII: direct and indirect. Direct PII can be used to directly link to a person's identity, which like name or social security number. The indirect PII, also known as \textit{quasi-identifier}, can reveal a person's identity when combined with other information, which like birthday, zip code, and sex~\cite{garfinkel2015identification}.

In order to share the data across different parties without leaking PII, there has been proposed many approaches to protect the individual's privacy, such as: \textit{de-identification}, \textit{synthetic data}, \textit{obfuscation}, and \textit{anonymization}. Below, we are going to give an overview of these methods and show their advancement and limitations.

\subsubsection{De-identification}

Traditionally, it is usually formulated the de-identification problem as a sequence labeling task. The very first attempt is to ask human annotators to label the PII tokens according to a pre-defined type list (such as Table~\ref{tab:pii_types}). However, it is unfortunately very costly and error-prone~\cite{douglass2004computer,douglass2005identification,neamatullah2008automated}. Later, the rule-based was proposed to detect and remove or substitute sentences or phrases that matched the pre-defined word patterns with regular expressions~\cite{dalianis2019pseudonymisation}. While rule-based systems  are relatively easy to create, the lack of generalizability make it hard to adapt to even the modest input data drifting.

In 2015, a rule-based and machine learning hybrid method was proposed, which utilize Conditional Random Field (CRF) model along with handcrafted regular expression rules to capture PII tokens, archiving around 91\% recall rate~\cite{yang2015automatic}, hence became the winner of 2014 i2b2 challenge \cite{stubbs2015automated}. Later,~\citet{dernoncourt2017identification} was the first attempt to combine Conditional Random Field (CRF) and bidirectional LSTMs to predict the PII tokens, with 97.84\% recall rate. \citet{ahmed2020identification} proposed a supervised self-attention based model to remove PII from Electronic Health Record (EHR), achieve 98.41\% recall rate on i2b2~\cite{stubbs2015automated} dataset. \citet{anjum2021identification} proposed an encoder-decoder architecture which consist of multi-head self-attention, modeling the de-identification problem as a sequence to sequence task instead of classification task. They archive 98.91\% recall rate on i2b2 dataset.

Recently, due to the advancement of Large Language Model (LLM), \citet{liu2023deid} proposed a zero-shot prompting based method, by utilizing OpenAI's GPT-4 model~\cite{openai2023gpt4}, can archive 99\% accuracy. Although the high accuracy and no needs of training data, its performance heavily dependents on the carefully designed prompt. Also, currently, most of LLMs are hosted by big companies, we need first upload sensitive data to those companies to get the de-identified result, it may compromise individual's privacy.

Despite the above language model based methods are very good at removing PII tokens, they need to follow some kind of predefined categories. For example, the U.S. Health Insurance and Accountability Act (HIPAA)~\cite{annas2003hipaa} defined 18 categories (see Table~\ref{tab:pii_types}) of re-identifying information need to be satinized before dissemination. In our cases, however, the re-identifiable information may not restrict to the predefined type of  identifier. For example: personal related events, achievements, awards, publications, rare diseases or special treatments, ...etc. By combining these personal sensitive aspects, it is very possible to identify the individuals, but hard to give a comprehensive list of predefined categories  due to its diverse nature.

To avoid the needs of predefine PII types, 
\citet{morris2022unsupervised} proposed an unsupervised de-identified framework, which can de-identify documents without using any human label. They use a pre-trained BERT style model to do the masked word prediction as a re-identification model, and then greedy masking words that minimized the re-identification probability. However, the resulted de-identified texts are less meaningful, because it may remove useful information. Also, we have no idea that what types of information would be removed, making this de-identified texts less utility to the downstream tasks.

\subsubsection{Synthetic Data}

Instead of removing PII tokens, another way is to generate a synthesized dataset, which mimic the distribution of real data with fake information, hence can be released to the public without privacy leakage.

~\citet{lee2018natural} was the first to propose a LSTM based synthetic generation model. They used the de-identified dataset to train their model. Although the data was already de-identified, there might still contain some residual PII tokens hadn't been removed, hence can be learned by the model and appear in the generated text. They claim that their model can automatically remove residual PII tokens due to the natural of model tend to sample high frequency terms, removing low-frequency terms (the PII tokens). However, they didn't provide any theoretical privacy guarantee. It is possible to recover the sensitive training data from neural network model through \textit{Membership Inference Attack}~\cite{shokri2017membership} or \textit{Unintended Memorization}~\cite{236216}, hence the privacy leakage level still dependents on the quality of de-identification method used to preprocess the training data.~\citet{melamud2019towards} proposed a LSTM based model with dropout regularization method as a randomness source to avoid private training data being leaked.~\citet{al2021differentially} proposed a decoder only, GPT-2~\cite{radford2019language} like model, along with \textit{Differentially Private SGD} (DP-SGD) algorithm~\cite{abadi2016deep}, which introduce randomness to the gradient during backpropagation to guarantee that privacy information leakage level satisfy \textit{($\epsilon$, $\delta$)-differential privacy}. 

While Differentially Privacy~\cite{dwork2006differential} theoretically preserves privacy at a given level, the utility of synthesized data heavily influenced by the privacy level. The higher the privacy level, the lower the utility of synthesized data. Also, most of the generation model needs large amount of training data to be able to produce realistic synthesized data. However, in the privacy-aware domain, we usually can only collect limited size of data, making the training of the generation model from scratch more challenging.

\subsubsection{Obfuscation}

The obfuscation method usually involves replacing the original content with symbols or texts that are different to the original content, but still can be processed as regular data by third parties without exposing original intent. After processed, only the data owner has the key to un-obfuscate the processed data to retrieve the processed information.
~\citet{reddy2016obfuscating} proposed a lexical substitution method, which modify the text by replace words with other one to obfuscate author's gender, while preserve text's fluency and meaning.
\citet{marujo2015privacy} proposed a secure multi-document summarization method, which utilize Secure Binary Embedding (SBE) hashes to obfuscate the content of document summarization from being accessed but can be processed by the third party, only the sender can retrieve the  processed passage from hashed representation, hence preserve the sender's privacy. \citet{hu2019obfuscation} proposed a syntactic aware text obfuscation model, which replace words in the original sentence while keep its syntactic structure the same, hence hide the original intent. 

Although the above methods can protect privacy by replacing words or transform text into some kind of symbols, they mainly focus on how to hide private information from other parties to know, not on publishing data for public use.

\begin{figure*}[htbp]
\centerline{\includegraphics[width=1\textwidth]{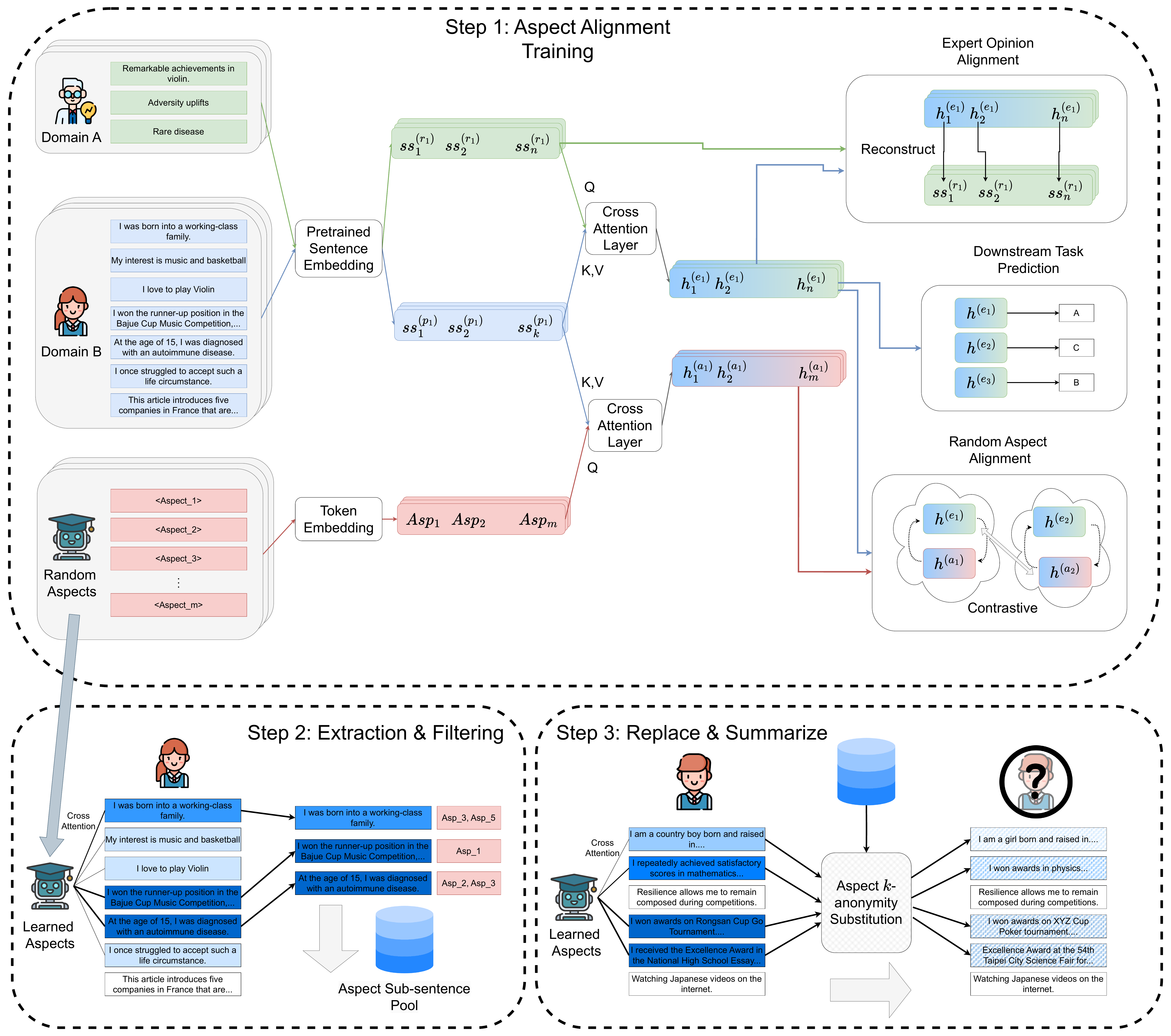}}
\caption{The Proposed Framework of \textsc{AspirinSum}}
\label{fig:framework_overview}
\end{figure*}

\subsubsection{Anonymization}

The first widely accepted model for Privacy-Preserving Data Publishing (PPDP) is called $k$-anonymity~\cite{samarati2001protecting, sweeney2002k}. In this model, a dataset achieves $k$-anonymity when each combination of indirect identifier values appears in at least $k$ records. When $k$ is greater than 1, it prevents re-identification, thus stopping identity disclosure. Most PPDP studies focus on structured databases, assuming it contains records detailing individuals' attributes. Despite this assumption, there are still few efforts tried to adapt $k$-anonymity to the unstructured text data~\cite{lison2021anonymisation}.

$k$-Safety~\cite{chakaravarthy2008efficient} redacted entity $e$ that does not satisfy at least $K-1$ other entities which have the same context words as $e$. However, it requires exhausted list possible context words for all entities, which is intractable. Also, it can only apply to a fixed set of sensitive entities, not used to be detecting unlisted sensitive entities.

\textit{C}-Sanitise~\cite{sanchez2016c} states that given a document $d$, the attacker is able to disclose sensitive term $t$ from $d$ by exploiting background knowledge $K$. Consider $K$ is the Web, $C$ is a set of entities to be protected, this approach can automatically detect sensitive terms $t$ through calculating the pointwise mutual information~\cite{anandan2011significance} between term $t$ and any entities in $C$ measured from their probability of (co)-occurrence in $K$. However, computing term co-occurrence across a large knowledge base is expensive, and the term-based calculation may overlook its contextual meaning.

In this work, by leveraging the multi-perspective natural of experts' comments, we can efficiently identify individual's sensitive aspects. Our goal is to make each aspect-related sub-sentence $k$-anonymity to prevent re-identification, hence more efficient than the term co-occurrence method.

\section{Proposed Methodology}
\label{sec:proposed_methodology}

\subsection{Overview}
\label{subsec:overview}

In this section, we present the individual components of our proposed framework, coined as \textsc{Asp}ect-based ut\textsc{i}lity-p\textsc{r}eserved de-\textsc{i}de\textsc{n}tification \textsc{Sum}marization framework (\textsc{AspirinSum}). \textsc{AspirinSum} is the framework that can extract aspect-related sub-sentences and substitute them with similar aspect sub-sentences from other persons' document to remove their linkages to the original person. The goal of the proposed \textsc{AspirinSum} is to learn a set of aspect token embeddings from the expert's opinions, and utilize these aspect tokens as a query to retrieve corresponding aspect sub-sentences from the unseen sensitive document. After retrieve, we use previous built aspect sub-sentence pool as a source of aspect sub-sentences to replace the original retrieved sub-sentences, ensure the sub-sentences comes from at least $k-1$ persons that have the similar aspect to avoid the original document owner's PSA been leaked. 

With the help of \textsc{AspirinSum}, one can de-identify documents without using a pre-defined type list and additional human labeling efforts. A high-level overview of the \textsc{AspirinSum} and its main components is provided in Figure~\ref{fig:framework_overview}. Below is a brief outline of the proposed framework in three steps:

\noindent \textbf{Step 1} \textit{Aspect Alignment Training} First, we propose \textbf{Expert Aspect Alignment (EAA)} mechanism (Figure~\ref{fig:model_detail}(a)), which can learn a set of aspect tokens from expert's reference notes and downstream task labels without using a predefined type list. The underlying model, called XAlign, is utilized to learn the relevance between expert's opinions and personal sensitive documents. We can use the learned aspect tokens to calculate the aspect relevance score for each sub-sentence appearing in the personal sensitive documents.

\noindent \textbf{Step 2} \textit{Extraction \& Filtering} In this step, we are doing \textbf{Aspect Sub-sentence Extraction (ASE)} (Figure~\ref{fig:model_detail}(b)). We use aspect tokens which learned in the step 1 as query input, and the training set of personal sensitive documents as key/value input of the XAlign model, to obtain the cross attention score and the predicted downstream task label for each document. By choosing a proper threshold, we can extract aspect-related sub-sentences from the cross attention score. Because the extracted sub-sentences still contain many non-relevant sub-sentences, we propose \textbf{Aspect Relevant Common Sequence Selection (ARCSS)} mechanism (Figure~\ref{fig:ARCSS_detail}), which utilized Longest Common Subsequence (LCS) metric~\cite{bakkelund2009lcs} to calculate the literal similarity between expert's reference notes and sub-sentences of the personal sensitive documents. Combining aspect relevance score and LCS metric, we select most non-relevant sub-sentences, train a relevant/non-relevant text classifier and apply it to the personal sensitive documents to filter out non-relevance texts. By iteratively apply above processes, the remaining texts will be more and more relevant to the expert's opinions. After filtering, we build up an aspect sub-sentence pool, which contains sub-sentence extracted from the training set documents, with columns document ID, downstream task class label, and aspect labels, to form an indexed database.

\noindent \textbf{Step 3} \textit{Replace \& Summarize} With the aspect sub-sentence stored in the aspect sub-sentence pool, we can use it to replace the sensitive sub-sentence of the input sensitive document. We first use learned aspect tokens to find where are the most sensitive aspects sub-sentences, and then send it into \textbf{Aspect $k$-anonymity Substitution (AKS)} (Figure~\ref{fig:substitution_detail}) module to replace with similar aspect sub-sentence from the sub-sentence pool. To ensure the replaced result satisfy $k$-anonymity, the candidate sub-sentences are comes from at least k persons. The AKS first search sub-sentence with most similar aspect by calculating \textit{aspect distance}. The aspect distance is the hamming distance between two sub-sentences' cross attention score after threshold. When the resulted sub-sentence comes from less than $k-1$ person, it increases distance by 1 to include more person. Until the resulted person greater equal than $k-1$, it samples a sub-sentence from the resulted sub-sentences, and use the sampled one to substitute the original sub-sentence.

In order to evaluate \textsc{AspirinSum}'s performance, we are not only consider its sub-sentence retrieval accuracy, but also consider its utility, fidelity and privacy of the generated summary. For utility, by comparing the downstream task classification accuracy trained on different method generated summaries, we can show that our method can preserve more utility than others. For fidelity, we investigate "trained on original and test on generated" setup, and the unsupervised clustering setup to demonstrate the agreement between original dataset and generated dataset. For privacy, we try to do the re-identification attack and the membership inference attack to show that our method can achieve high privacy level.

In following sections, we will elaborate each step with more details.

\subsection{Preliminaries}
\label{subsec:preliminaries}

\subsubsection{Cross Attention}

Cross attention was first appearing in the original Transformer architecture~\cite{vaswani2017attention}. In the encoder-decoder architecture, it utilized to mix encoder's information into the decoder's domain. For example, to translate source language into target language, it involves mixing the source information into the target domain.

\citet{rombach2022high} used cross attention in their Latent Diffusion Model to mixing the conditioning text information into the image generation process. \citet{xiao2022attend} proposed a content-controllable summarization method, by manipulating the cross attention score to more relevant to the controlling aspect
token, it can generate summaries more related to the given controlling aspect. \citet{bansal2024llm} proposed a parameter efficient fine-tuning module called CALM, which use cross attention layer to bridge two language models, the anchor LLM and the augmenting LLM, to acquire new combined skills of the two model without fine-tuning each of them.

In this work, we propose \textbf{Expert Opinion Learning (EOL)} mechanism, which leveraging the cross attention mechanism to create the connection between expert's comments and the personal sensitive document by treating expert's comment as query and sensitive document as key/value. After training, the connection was established. We can utilize the resulted cross attention score to identify which part in the document is most relevant to the query. Hence, can be used to extract sensitive aspects.

\subsubsection{Contrastive Learning}

The fundamental concept of Contrastive Learning (CL) involves taking pairs of similar data samples as positive pairs and dissimilar data samples as negative pairs. The objective is to train the model to produce representations of samples where the distance between representations is minimized for positive pairs and maximized for negative pairs. The challenge lies in obtaining positive pairs without human annotation. For visual data, SimCLR~\cite{chen2020simple} has been the most successful attempt at applying CL to learn effective visual representations. By applying various image augmentation techniques such as translation, rotation, and cropping, similar yet distinct images can be generated as positive pairs without human annotation.

However, for textual data, automatically augmenting text without altering its meaning is not straightforward. To address this, SimCSE~\cite{gao2021simcse} proposed a different approach: applying a dropout mask to the embedding of the same sentence twice to create a positive pair. Essentially, the dropout operation generates two different views of the same input data.

In this work, we proposed \textbf{Random Aspect Alignment (RAA)}, which randomly select aspect tokens as query to the same document, to create different views of the same document from different aspects. Each random aspect query to the same document should be similar, to the different document should be dissimilar, hence can be a positive/negative pair. We use random aspect tokens and expert's comments as queries to the same document respectively, treat them as a positive pair, and negative pair for query to the different document, then applying CL to align random aspect with the expert's comments.

\begin{figure*}[ht]
\centerline{\includegraphics[width=1\textwidth]{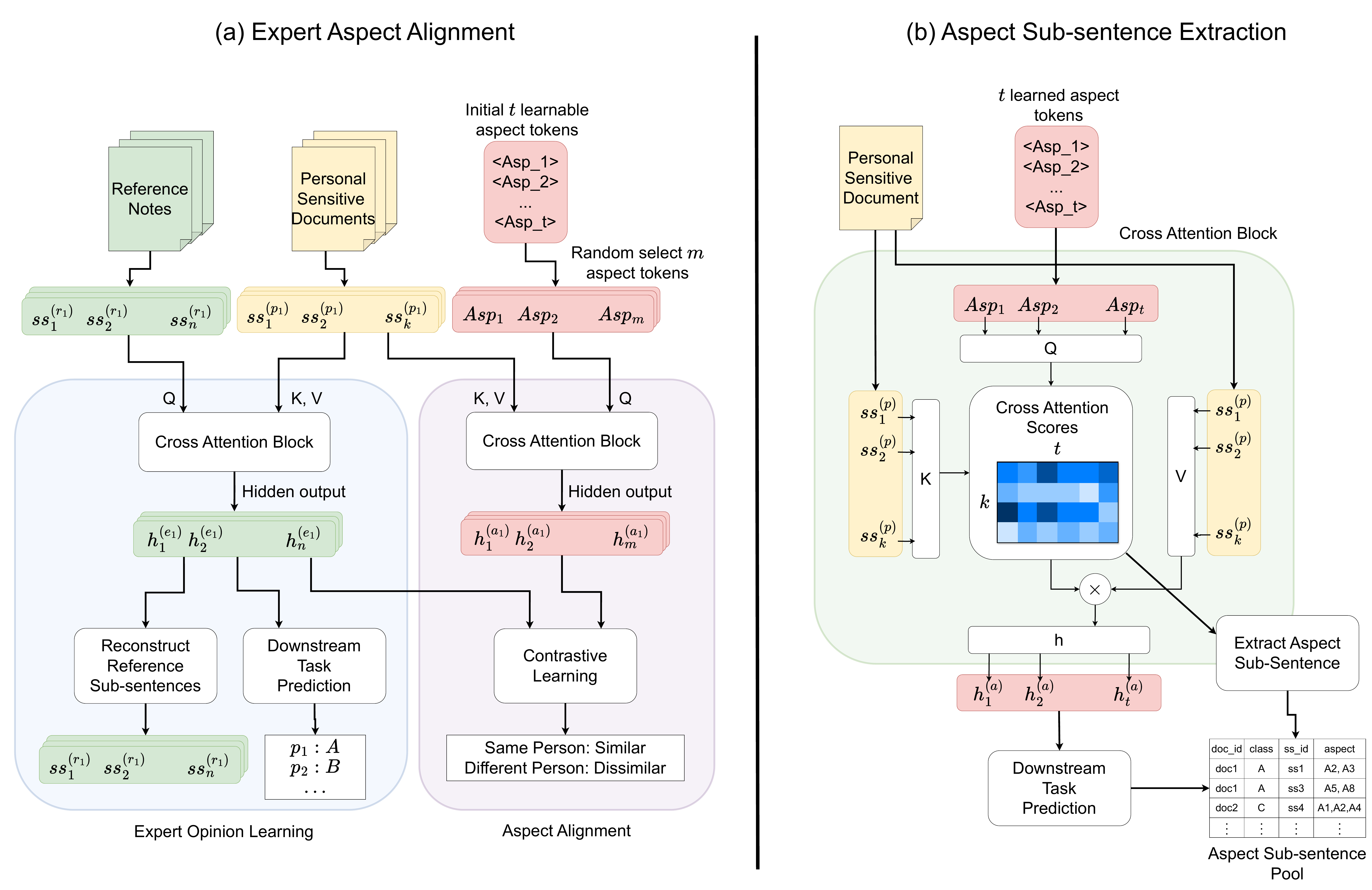}}
\caption{The details of the proposed XAlign model}
\label{fig:model_detail}
\end{figure*}

\subsubsection{$k$-anonymity}

One of the most famous anonymization methods is $k$-anonymity~\cite{sweeney2002k}. The idea is simple: conceal an individual's information within a group of size $k$, ensuring that the information of the remaining $k-1$ individuals is indistinguishable from the target individual. To achieve this with tabular or relational structured data, we generalize each person's attribute values into nonspecific ones. For instance, instead of using a precise age (e.g., 25), we employ age ranges (e.g., 20-30) to represent the individual's age. If there were at least $k$ persons in this range, we say that the individual's age is hiding in the size $k$ group.

However, applying the generalization process to textual data is challenging. The most common de-identify approach involves identifying sensitive text and redacting it. Yet, this method often requires costly human labeling and risks information leakage due to imperfect redaction.

In this proposal, we introduce \textbf{Aspect $k$-anonymity Substitution}. Inspired by the aspect-based summarization, which can be utilized to extract aspect related terms. We use this idea to extract PSA from sensitive documents. We also build up an aspect sub-sentence pool. When there are sensitive documents to be de-identify, it first searches similar aspect sub-sentence from the pool, ensures the resulted sub-sentences comes from at least $k-1$ person, and sample a sub-sentence from the search result to replace the original sub-sentence. This ensures that each sensitive sub-sentence is hiding in the $k$ person group.

\begin{figure}[ht]
\centerline{\includegraphics[width=0.5\textwidth]{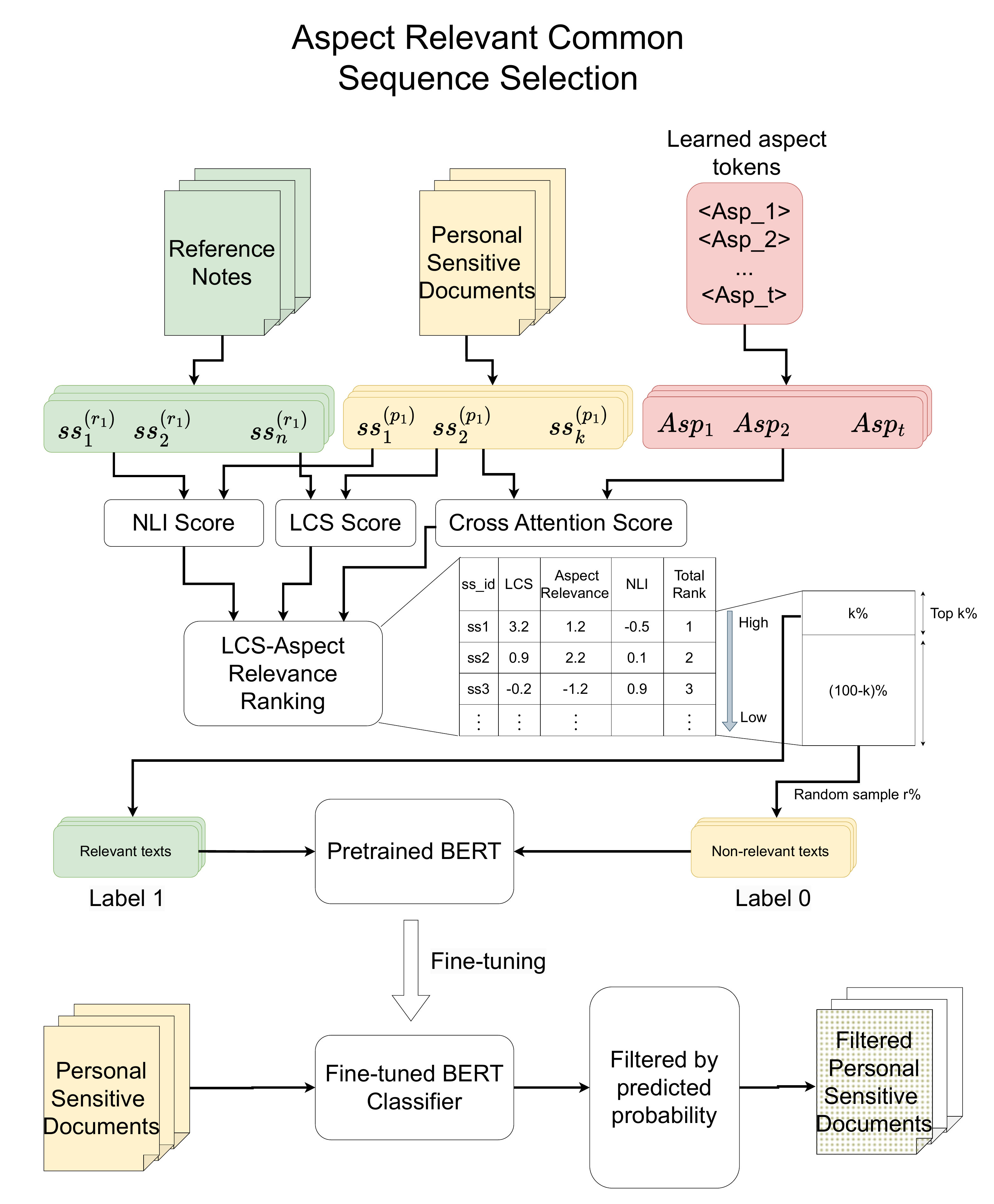}}
\caption{The details of the proposed Aspect Relevant Common Sequence Selection}
\label{fig:ARCSS_detail}
\end{figure}

\subsection{Problem Formulation}
\label{subsec:problem_formulation}

Let $C_R$ denote a corpus of expert's reference notes about persons (e.g. committee member's comments for students, doctor's clinical notes for patients or interviewer's notes for job applicants), $C_S$ denote a corpus of persons' sensitive documents (e.g. self-statement, medical records or resume), and $D_L$ denote a dataset of downstream task label about the evaluation of persons (e.g. student's academic performance grade, patient's disease severity level or applicant's job suitability level). Let $P=\{p_1, p_2,...\}$ denotes the set of persons, $S=\{s^{(p_1)}, s^{(p_2)}, ...\}$ denotes the set of sensitive documents owned by persons, $R_p=\{r^{(p)}_1, r^{(p)}_2, ..., r^{(p)}_{|E|}\}$ and $L_p=\{l^{(p)}_1, l^{(p)}_2, ...,l^{(p)}_{|E|}\}$ denotes the set of reference notes and downstream task labels given by experts for a person $p \in P$, where $E$ is the set of experts. For each reference note $r^{(p)}_e, e \in E$ and sensitive document $s^{(p)}$ can be split into a sequence of sub-sentences $SS_{r^{(p)}_e}=\{ss^{(p,e)}_1, ss^{(p,e)}_2,...\}$ and $SS_{s^{(p)}}=\{ss^{(p)}_1, ss^{(p)}_2, ...\}$ respectively.

The goal is to learn the cross attention score between $SS_{r^{(p)}_e}$ and $SS_{s^{(p)}}$ to represent the expert's opinions, and use a set of random initialized aspect tokens to align with it. At inference time, it can use aligned aspect tokens to calculate the cross attention score on the unseen sensitive document without giving expert reference notes.

\subsection{Expert Aspect
Alignment}
\label{subsec:expert_aspect_alignment}

In Figure~\ref{fig:model_detail}(a) left, we firstly train a cross attention model with expert's reference notes as query and personal sensitive document as key/value input to learn how the expert's opinions attend to the sensitive document. The Cross Attention Block is modified from the original Transformer architecture~\cite{vaswani2017attention}. Due to that fact that softmax activation may produce too sparse attention score across a long sequence, so instead of using softmax to calculate attention score, we use sigmoid function to calculate our Cross Attention Score (CAS):

$$
CAS(Q,K) := sigmoid(\frac{QK^T}{\tau\sqrt{d_k}})
$$

\noindent where $\tau$ is a scaling factor, and normalize CAS as the form of probability to calculate weighted sum of V:

$$
h(Q,K,V) = \frac{CAS(Q,K)}{\sum{CAS(Q,K)}}V
$$

\noindent In this way, we separate the normalization term from the attention score, resulted in a more meaningful attention score. Further, we add a dropout and batch norm layer to the hidden outputs to make the training process more stable.

The length of hidden outputs $\{h_1^{(e)}, h_2^{(e)}, ... h_n^{(e)}\}$ is the same as the input query sequence, and each hidden state can be seen as a weighted sum of the input value sequence. When we ask the model to reconstruct the query sequence from hidden states, we are actually forcing the model to attend to the parts of the value sequence that most relevant to the query sequence. Hence, the resulted cross attention score should reflect expert's opinions presented in the query sequence. The loss function of reconstruction objective is:

$$
\mathcal{L}_{reconstruct} = \frac{1}{n}\sum_{i=1}^{n} MSE(h_i^{(e)}, ss_i^{(r)})
$$

\noindent where $ss_i$ is the input embedding of the query sub-sentence.

Optionally, we can add auxiliary task such as downstream task prediction to acquire more relevance signal from labels, by simply averaging hidden states and passing it to the MLP classification layer. The loss function of auxiliary task is:

$$
\mathcal{L}_{auxiliary} = CrossEntropy(y, \hat{y})
$$

\noindent where $y \in L$ is the ground truth label, $\hat{y}=softmax(MLP(\frac{1}{n}\sum_{i=1}^{n}h_i^{(e)}))$ is the predicted probability.

In Figure~\ref{fig:model_detail}(a) right, we train the same model with the same key/value input but different query to align the expert's opinions. Imagine 
the total aspects used in the whole expert's opinions can be represented by $t$ tokens. For each expert, they only use $m$ tokens in their reference note each record. 

During training, we randomly initialize $t$ learnable aspect tokens $A=\{\text{\textless}asp\_1\text{\textgreater},$ $\text{\textless}asp\_2\text{\textgreater}, ..., \text{\textless}asp\_t\text{\textgreater}\}$, and randomly sample $m$ aspect tokens as the query sequence for each record. Because the input key/value is the same as the expert opinion learning part, the only difference is the query sequence. 
Although the query sequence from the two parts usually have different length, which the output hidden states can not be simply compared, we can still use the average of hidden states to take the comparison.

Because the average of hidden states is just the weighted sum of the value sequence, and because the two query sequences should both be relevant to the same input value sequence no matter how they difference, the resulted average hidden states should be as close as possible. On the other hand, if the input value sequences are different between the two parts, the resulted average hidden states should be as far as possible. The pairwise contrastive loss is:

$$
\ell_{i,j} = -log \frac{exp(sim(z_i, z_j)/\tau_c)}{\sum_{k=1}^{2N} \mathbb{1}_{k\neq i} exp(sim(z_i, z_k)/\tau_c)}
$$

\noindent where $z_k = \frac{1}{n}\sum_{i=1}^{n}h_i^{(e)}$ when $k \in [1,...,N]$, and $z_k = \frac{1}{m}\sum_{i=1}^{m}h_i^{(a)}$ when $k \in [N+1,...,2N]$, $N$ is batch size, $sim(\cdot)$ is cosine similarity, $\tau_c$ is temperature parameter. The total loss of alignment is:

$$
\mathcal{L}_{alignment} = \frac{1}{2N} \sum_{k=1}^{N} [\ell_{k,N+k} + \ell_{N+k,k} ]
$$

Therefore, leveraging contrastive learning, it pulls the similar and pushes the dissimilar between expert's opinions and random aspect tokens, the resulted aspect tokens should learn to align with expert's opinions. The total loss is:

$$
\mathcal{L} = \mathcal{L}_{reconstruct} + a \mathcal{L}_{auxiliary} + b \mathcal{L}_{alignment}
$$

\noindent where $a, b$ is the weighting between each loss.

Hence, we can use the aligned aspect tokens to represent the expert's opinions, and use it as the query sequence during inference.

\begin{figure*}[ht]
\centerline{\includegraphics[width=0.7\textwidth]{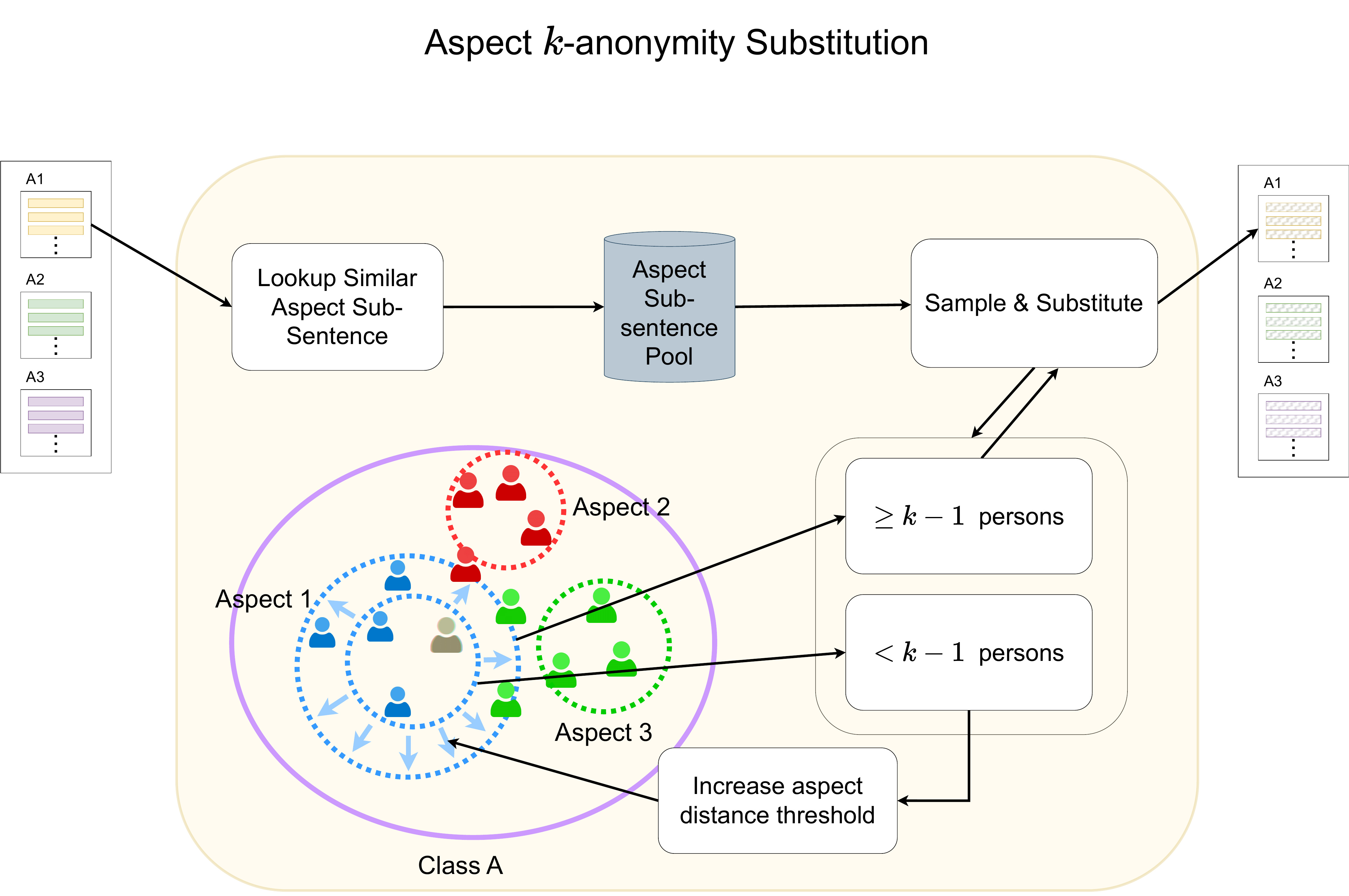}}
\caption{The details of the Aspect $k$-anonymity Substitution}
\label{fig:substitution_detail}
\end{figure*}

\subsection{Aspect Sub-sentence Extraction}
\label{subsec:aspect_sub_sentence_extraction}

In Figure~\ref{fig:model_detail}(b), during inference, we take all $t$ aligned aspect tokens in $A$ as query sequence and a personal sensitive document $s^{(p)}$ as key/value sequence to input to the XAlign model. The resulted CAS matrix then used to extract aspect sub-sentences from the sensitive document. Before extraction, we first perform standardization along sequence length axis on CAS:

$$
\overline{CAS} := \frac{CAS - mean(CAS)}{std(CAS)}
$$

After standardization, we binarize $\overline{CAS}$ with a given \textit{standard attention threshold} $\alpha$:

$$
{bin(\overline{CAS})} : = 
\left\{ 
\begin{array}{ll}
1, & if\ a_i \geq \alpha \\ 
0, & if\ a_i < \alpha 
\end{array} 
\right.
$$

\noindent where $a_i$ is the element of the $\overline{CAS}$ matrix.

The size of the $bin(\overline{CAS})$ is $k \times t$, where $k$ is the length of key/value, and $t$ is the length of query. Because there are $t$ aspects, in order to determine whether to extract sub-sentence or not, we further aggregate aspects into a single value by summing over aspects and define another threshold called \textit{aspect consensus number} $\beta$. The sub-sentence Extraction Mask can be defined as the following:

$$
{Extraction\ Mask} : = 
\left\{ 
\begin{array}{ll}
1, & if\ \sum_{i \in t}{b_i} \geq \beta \\
0, & if\ \sum_{i \in t}{b_i} < \beta 
\end{array} 
\right.
$$

\noindent where $0 \leq \beta \leq t$, and $b_i$ is the element of the $bin(\overline{CAS})$ matrix.

Optionally, if the XAlign model had been trained with auxiliary downstream task prediction, we can obtain the predicted downstream task label and CAS simultaneously.

By repeating this process for all sensitive documents in $C_S$, we can establish an aspect sub-sentence pool $T$, which composed of predicted downstream label $l$ and aspect labels $b_i$ for each extracted sub-sentence.

\subsection{Aspect Relevant Common Sequence Selection}
\label{subsec:aspect_relevant_common_sequence_selection}
In order to filter out texts that unrelated to the expert's opinions from personal sensitive documents, we purpose to train a classifier to differentiate relevant and non-relevant texts. Figure~\ref{fig:ARCSS_detail} shows that how we select relevant and non-relevant training samples. The relevant one is simple, we directly use the expert's reference notes as the source of relevant texts. To select non-relevant texts, we first utilize the learned aspect tokens to calculate the \textbf{Aspect Relevance Score (ARS)} for each sub-sentence in a personal sensitive document. The ARS for the $j$-th sub-sentence is defined as:

$$
ARS(ss_j) := \frac{1}{|Aspects|} \sum_{i \in \{Aspects\}} CAS_{i, j}
$$

After obtain ARS, we rank each sub-sentence from high to low, e.g. the rank-1 sub-sentence should have the highest ARS among all sub-sentences in the same personal sensitive document, and the rank-2 have the second-high ARS, and so on.

Another score we need to consider is called \textbf{Longest Common Subsequence Similarity (LCSS)}, following the definition of the LCS metric~\cite{bakkelund2009lcs}, LCSS can be defined as:

$$
LCSS := \frac{|LCS(s1, s2)|}{max(|s1|, |s2|)}
$$

\noindent where $s1$, $s2$ are any given text sequences. When two text sequences are literally similar, i.e. they share a long common subsequence, the LCSS will be close to 1.

To calculate LCSS for each sub-sentence, we first concatenate all expert's reference notes which gives to the same person as a single reference sequence, and calculate LCSS between it and each sub-sentence which comes from the personal sensitive document of the same person. After obtain LCSS, we also rank each sub-sentence based on LCSS, the same as ARS ranking do. 

The use of literal similarity is curial for domain specific data, because the expert tends to use special terms or domain specific abbreviations that didn't appear in other places, these terms usually can not be well represented by the pretrained sentence encoder.

The total rank is simply adding the rank value of ARS and LCSS, e.g. sub-sentence 1 has ARS rank 5 and LCSS rank 2, its total rank is 5+2=7. After obtain the total rank, we then sort each sub-sentence by its total rank. 

We keep the top k ranked sub-sentence untouched, and select the last 20\% of the top N-k sub-sentences as the non-relevant training samples. The reason we keep top k sub-sentence untouched is that it has high agreement between semantic and literal in both high similarity. Similarly, we select the last 20\% as non-relevant samples is because it has high agreement between semantic and literal in both low similarity. The middle 80\% may contain more noise due to the disagreement between semantic and literal similarity. 

With the relevant and non-relevant samples, we are able to train a text classifier, by fine-tuning the pretrained BERT. After fine-tuning, given a personal sensitive document, we can rank each sub-sentence by its predicted probability of the relevant class, keep top k untouched, and remove last 25\% sub-sentence. Note that, we remove a bit more (5\%) than we selected as non-relevant samples due to the model's ability to generalize to the unseen data (5\% away from training sample might be safe).

\subsection{Aspect $k$-anonymity Substitution}
\label{subsec:aspect_based_k_anonymity_substitution}

As shown in Figure~\ref{fig:substitution_detail}, to get the de-identified summary of a given sensitive document $s^{(p)}$, we first extract aspect sub-sentences from the document through Aspect Sub-sentence Extraction process. For each extracted sub-sentence, searching the pool $T$ to find a set of sub-sentences ${SS}^{(Q)}=\{ss^{(Q)}_1, ss^{(Q)}_2, ...\}$ that have aspect labels similar to the original one, randomly sample from it and substitute the original sub-sentence with the sampled one. Note that, $Q$ is the set of persons that have the same downstream task label as the given sensitive document excluding person $p$, and make ensure that $|Q| \geq k-1$ to satisfy $k$-anonymity.

Because aspect labels are already binarized, the aspect distance between two sub-sentences is simply the hamming distance of aspect labels. When $|Q| < k-1 $, it will increase the aspect distance by 1 and search pool $T$ again to include more person. Repeating this process until $|Q| \geq k-1$.

Optionally, if the downstream task label presents, it can further narrow down the search space by selecting only the sub-sentences from documents that have the same downstream task label class as the original document.


\begin{table*}[ht]
\centering
\begin{tabular}{lcccc}
\hline
\textbf{Method} & \textbf{Extraction ratio} & \textbf{P} & \textbf{R} & \textbf{F1} \\
\hline
human label & 0.06 & - & - & - \\
\hhline{=====}
random 0.2 & 0.2 & 0.052 & 0.177 & 0.074 \\
hdbscan (min\_cluster\_size=15, th=0.75) & 0.2 & 0.059 & 0.224 & 0.086 \\
hdbscan (min\_cluster\_size=30, th=0.87) & 0.2 & 0.062 & 0.218 & 0.088 \\
hdbscan (min\_cluster\_size=60, th=0.99) & 0.22 & 0.067 & 0.268 & 0.100 \\
XAlign(epoch=150, $\alpha=1$, $\beta=5$) & 0.19 & 0.126 & \textbf{0.455} & 0.182 \\
XAlign(epoch=150, $\alpha=1.7$, $\beta=5$) & 0.09 & 0.148 & 0.261 & 0.167 \\
XAlign+ARCSS(threshold=0.1, iter=1) & 0.14 & 0.180 & 0.445 & 0.219 \\
XAlign+ARCSS(threshold=0.5, iter=1) & 0.05 & 0.280 & 0.265 & \textbf{0.220} \\
XAlign+ARCSS(threshold=0.7, iter=1) & \textbf{0.03} & \textbf{0.336} & 0.175 & 0.180 \\
\hline
\end{tabular}
\caption{The Precision/Recall/F1 score of the extracted sub-sentence on testing set}
\label{tab:human_labeled_test_set}
\end{table*}


\section{Preliminary Results}
\label{sec:preliminary_results}

\subsection{Dataset}
\label{subsec:dataset}

To validate the applicability of our method in real world personal sensitive document data, we conduct experiments on our newly created dataset, called \textbf{High-School Student's College Application (HSSCA)}. The origin of HSSCA is National Tsing Hwa University Center for Admission and Strategy (NTHU-CAS). Every year, many high-school students will submit their application to the NTHU-CAS, the application contains self-statement, recommendation letters and academic transcript data. After receiving the students' applications, the university will assign 4-5 committee members to grade each student and leave some comment for later reference. The data was collected from 2017 to 2021, excluding 2018 due to the presence of mismatch of committee members' comments and student's self-statement in the original data. The total number of the student application in original HSSCA dataset is 1789, after removing 51 cases from 2018, it results total 1738 cases. We split the original dataset $C_S$ into training/testing set $C_S^{(train)}/C_S^{(test)}$. The number of instances in the training/testing sets are 1389/349 respectively.

In this preliminary study, we consider the student's self-statement as the sensitive document data, committee members' comments as the reference notes data, and the grade they gave as the downstream task label data. The grade were initially given in numerical score, ranging from 65 to 100, and can be converted into 4 classes \{A, B, C, F\} through a predefined class-score mapping: \{A: above 91, B: 81-90, C: 71-80, F: below 70\}. We can obtain a single final grade for each student by simply averaging grade score from all committee members. Note that, we take each committee member's comment and grade as an independent instance during training, and merging them into a single comment and grade for each student during evaluation.

Before sending documents into the \textsc{AspirinSum} framework, we first split documents into sub-sentence chunks by punctuation. To avoid resulted sub-sentences too short, we 
apply a merge process to each sub-sentence. Leveraging BERT's Next Sentence Prediction (NSP) label, we concatenate the current sub-sentence with the next sub-sentence when NSP label is 1, and keep it untouched when NSP label is 0. After the sub-sentence merge process, because our data contains mixing of Chinese and English texts, we use multilingual SBERT model \textit{sentence-transformers/paraphrase-multilingual-MiniLM-L12-v2}~\cite{reimers-2019-sentence-bert,reimers-2020-multilingual-sentence-bert} to convert sub-sentences into embeddings with dimension 384. We use the resulted sub-sentence embeddings to do the subsequent process. Also, we manually labeled self-statement sub-sentences that matching the corresponding committee member's comments in testing set for evaluation purpose.

\subsection{Experiment on Expert Aspect Alignment}
\label{subsec:experiment_on_expert_aspect_alignment}

We first investigate the model's ability to extract the sub-sentences which corresponding to the committee member's comments from the student's self-statement. The purpose of this experiment is to show that how well the model learned aspect tokens align with expert's reference notes (Sec~\ref{subsec:expert_aspect_alignment}). Since the expert's reference notes are comments on the given sensitive document, each expert's notes should have its corresponding sub-sentence appear in the given document. By evaluating how well the model extracted sub-sentences align with human labeled sub-sentences, we can show that if the learned aspect tokens really align with expert's aspects.

\subsubsection{Setup}
In the random baseline, we randomly sample 20\% of sub-sentences from the document, resulted in the lowest precision, recall and f1 score. In HDBSCAN baseline, we conduct three experiments, with hyperparameters min\_cluster\_size=\{15, 30, 60\} and threshold=\{0.75, 0.87, 0.99\} respectively. The min\_cluster\_size will affect the number of clusters, and the threshold determines whether the data should be treated as an outlier. The higher the min\_cluster\_size, the less the cluster numbers; the higher the threshold, the more data to be considered as outlier. We use the sub-sentence embedding of committee member's comments to train the HDBSCAN model to obtain clusters of expert's opinions, and inference on the given sensitive documents. If the model assign any cluster to the sub-sentence, we treat it as extracted; if the model assign -1 to the sub-sentence, it means the sub-sentence is an outlier, we treat it as not extracted.

During training, we choose $t=10$ as total aspects tokens and $m=5$ as random aspect number. We set the scaling factor $\tau=0.007$ in the sigmoid function, $\tau_c =0.5$ in the contrastive loss, dropout probability 0.7, and $b=0.01$ is the weight of alignment loss. We use Adam optimizer with learning rate $10^{-4}$ and weight decay 0.015, running 150 epochs with batch size 64 to obtain the best result on the testing set. Note that, we did not use any labeled data during training, the labeled data only utilized for evaluation purpose.

When inference, we extract the CAS, binarized it with $\alpha = 1.0$ and calculate its Extraction Mask by setting $\beta = 5$. The reason for $\alpha = 1.0$ is simply choosing standardized attention scores larger than one standard deviation, and for $\beta = 5$ is because we use total $t=10$ aspects tokens for training, and the half of total should be a reasonable number to achieve consensus.

\begin{table*}[ht]
\centering
\resizebox{\textwidth}{!}{%
\begin{tabular}{lccccccc}
\hline
\textbf{} & \textbf{Acc.} & \textbf{P} & \textbf{R} & \textbf{F1} & \textbf{P(weighted)} & \textbf{R(weighted)} & \textbf{F1(weighted)} \\ \hline
train on original, test on original & 0.458 & 0.298 & 0.283 & 0.260 & 0.407 & 0.458 & 0.396 \\ \hhline{========}
train on random-substitute, test on original & 0.458 & 0.332 & 0.336 & 0.329 & 0.438 & 0.458 & 0.441 \\ 
train on aspect-k-anonymity, test on original & 0.458 & 0.322 & 0.306 & 0.298 & 0.432 & 0.458 & 0.422 \\ \hline
train on original, test on random-substitute & \textbf{0.461} & 0.226 & 0.265 & 0.210 & 0.347 & \textbf{0.461} & 0.350 \\ 
train on original, test on aspect-k-anonymity & 0.453 & \textbf{0.379} & \textbf{0.279} & \textbf{0.255} & \textbf{0.463} & 0.453 & \textbf{0.384} \\ \hline
\end{tabular}%
}
\caption{Downstream Task Utility (classification)}
\label{tab:downstream-task-utility}
\end{table*}

\begin{table*}[h]
\centering
\resizebox{\textwidth}{!}{%
\begin{tabular}{lccccccc}
\hline
\textbf{} & \textbf{Acc.} & \textbf{P} & \textbf{R} & \textbf{F1} & \textbf{P(weighted)} & \textbf{R(weighted)} & \textbf{F1(weighted)} \\ \hline
train on original, test on random-substitute & 0.198 & 0.172 & 0.135 & 0.096 & 0.184 & 0.198 & 0.123 \\ 
train on original, test on aspect-k-anonymity & \textbf{0.218} & \textbf{0.200} & \textbf{0.175} & \textbf{0.165} & \textbf{0.209} & \textbf{0.218} & \textbf{0.188} \\
\hline
train on random-substitute, test on original & 0.106 & 0.113 & 0.126 & 0.103 & 0.132 & 0.106 & 0.107 \\ 
train on aspect-k-anonymity, test on original & \textbf{0.198} & \textbf{0.216} & \textbf{0.196} & \textbf{0.191} & \textbf{0.234} & \textbf{0.198} & \textbf{0.203} \\ \hline
\end{tabular}%
}
\caption{Document fidelity (clustering agreement) with K-Means (k=8)}
\label{tab:document-fidelity-K-Means-k-8}
\end{table*}

\begin{table}[h]
\centering
\begin{tabular}{lcc}
\hline
\textbf{} & \textbf{ARI} & \textbf{AMI} \\ \hline
original vs. random-substitute & 0.002 & 0.001 \\ \hline
original vs. aspect-k-anonymity & \textbf{0.012} & \textbf{0.016} \\ \hline
\end{tabular}
\caption{ARI and AMI with K-Means (k=8)}
\label{tab:ARI-AMI-K-Means-k-8}
\end{table}

\subsubsection{Results}

Table~\ref{tab:human_labeled_test_set} shows the average Precision, Recall and F1 score of the extracted sub-sentences on the testing set for different method. We use human label as ground truth. As we can see, human label gives the lowest extraction ratio 0.06, which means experts usually take only 6\% of information from the document to summarize a person. The best F1 score that HDBSCAN is 0.1, however, our XAlign method achieves 0.182 F1 score, outperform the best HDBSCAN setting around 82\% with similar extraction ratio 0.19. By adjusting $\alpha$ to 1.7, we can reduce the extraction ratio to 0.09, approaching to the human label, with precision 0.148, outperform the best HDBSCAN precision around 120\%. By adding ARCSS process, it shows that there is a significant improvement in precision compare to the solely XAlign model in the similar recall setting. For example, comparing XAlign(epoch=150, $\alpha$ = 1.7, $\beta$ = 5) and XAlign+ARCSS(threshold=0.5, iter=1), the improvement is near 100\%.


Although the absolute precision looks quite low at first glance, our method tries to learn comprehensive aspects from all the expert's comment presented in the training set. Besides, the human label used in testing data are solely based on the available expert's notes, it means that the label may not reflect the diversity of the aspects, because the experts usually only write down the most significant part in their notes, the notes are not meant to be comprehensive. Hence, the label data may not as comprehensive as predicted label, make the current labeled testing set is not a good evaluation proxy. To see the extracted example, go to Appendix~\ref{app:example_extracted_sub_sentences}.



\begin{center}
\begin{table*}[h]
\centering
\begin{tabular}{lcccc}
\hline
\multicolumn{1}{c}{\textbf{}} & \multicolumn{4}{c}{\textbf{Re-identify Accuracy}} \\ \cline{2-5} 
\multicolumn{1}{c}{\textbf{}} & \textbf{Top-1} & \textbf{Top-5} & \textbf{Top-10} & \textbf{Top-100} \\ \hline
train on original, test on original & 0.99216 & 0.99888 & 0.99888 & 1.00000 \\ 
train on original, test on random-substitute & 0.00112 & 0.00448 & 0.00840 & 0.08343 \\ 
train on original, test on aspect-k-anonymity & 0.00112 & 0.00504 & 0.01176 & 0.08007 \\ \hline
\end{tabular}
\caption{Re-identify Accuracy (Lower is better)}
\label{tab:re-identify-accuracy}
\end{table*}
\end{center}

\subsection{Experiment on Downstream Task Utility and Document Fidelity}

\subsubsection{Utility}

To evaluate the utility of our de-identified summaries, we first follow the literature's setup~\cite{hu2023sok, harder2021dp, torkzadehmahani2019dp, chen2020gs}, training a classification model on the generated de-identified data, and testing on the original data to obtain its classification accuracy. 

To be more specific, we use documents in the training set $C_S^{(train)}$ to train our XAlign model, extracting and building up the aspect sub-sentence pool $T$, then apply Aspect $k$-anonymity Substitution on the training set itself to obtain the de-identified training set $C_{\bar{S}}^{(train)}$. 

We then use $C_{\bar{S}}^{(train)}$ and the original downstream task labels to train a classification model $\bar{\mathcal{M}}_{classify}$, and calculating its accuracy/precision/recall/f1 score on the original testing set $C_{S}^{(test)}$.




\subsubsection{Fidelity}

Different from utility, there is another concept called \textit{fidelity}, which usually refers to preserving statistic or structural properties of the original dataset. To evaluate fidelity, suppose you have the original document dataset and the corresponding de-identified document dataset. If these two datasets were similar in their statistics properties, we say the de-identified document dataset have high fidelity to the original document dataset.

Specifically, we use the original training set $C_S^{(train)}$ and the de-identified training set $C_{\bar{S}}^{(train)}$ to train the clustering model respectively, and the resulted model $\mathcal{M}_{cluster}$ and $\bar{\mathcal{M}}_{cluster}$ can then be applied to the original testing set $C_{S}^{(test)}$ and the de-identified testing set $C_{\bar{S}}^{(test)}$ respectively.

We firstly take the model $\mathcal{M}_{cluster}$ to apply to the $C_{S}^{(test)}$ to get the "train on original, test on original" cluster labels $L_{cluster}^ {(ori, ori)}$, and then apply to $C_{\bar{S}}^{(test)}$ to get the "train on original, test on de-identified" cluster labels $L_{cluster}^{(ori, deid)}$. We use $L_{cluster}^{(ori, ori)}$ as ground truth labels and $L_{cluster}^{(ori, deid)}$ as predicted labels to calculate its accuracy/precision/recall/f1 score.

On the other hand, in order to match the utility setting, we also perform "train on de-identified, test on original" setup. We take the model $\bar{\mathcal{M}}_{cluster}$ to apply to the $C_{S}^{(test)}$ to obtain cluster labels $L_{cluster}^ {(deid, ori)}$ as ground truth, and apply to $C_{\bar{S}}^{(test)}$ to obtain cluster labels $L_{cluster}^{(deid, deid)}$ as predicted label, then calculate its accuracy/precision/recall/f1 score.

Furthermore, we can evaluate the degree of clustering agreement by calculating Adjusted Rand Index (ARI) and Adjusted Mutual Information (AMI) between "train on original" $L_{cluster}^ {(ori)}$ and "train on de-identified" $L_{cluster}^ {(deid)}$ without using testing set.

\subsubsection{Setup}

Inspired by Inception Score (IS)~\cite{li2017alice} and Frechet Inception Distance (FID)~\cite{heusel2017gans}, which utilize a pretrained large image classification model to obtain the classification probability or encoded features of the input images  and calculate the difference between real and generated images, we use the pretrained SBERT model to generate the sub-sentence embeddings. However, different from image, the classification of text data usually depends on how human perceive it. So, in order to calculate the difference between real and generated text, we either need further fine-tuning the language model on the downstream classification task or use embeddings to train the clustering model.

In the utility setting, we simply take the average of the sub-sentence embeddings and passing it into the classification model. We use Xgboost with hyperparameters n\_estimators=100, max\_depth=3, learning\_rate=0.1, objective='multi:softprob', num\_class=4, eval\_metric='mlogloss' and early\_stopping\_rounds=10 as our classification model.

In the fidelity setting, we use Kmeans~\cite{hartigan1979algorithm} with k=8 cluster as our clustering model.

To de-identify documents, we choose $k=5$ persons in the Aspect $k$-anonymity Substitution.

\subsubsection{Results}

Table~\ref{tab:downstream-task-utility} shows the utility of the de-identified summaries on the downstream task. The first row "train on original, test on original" means using original document without any de-identification to train and test the classification model. This setting is also considered as the normal training process, the performance should be higher than the de-identified setting, so we treat it as our best case performance baseline. The second row "train on random-substitute" means that the aspect extracted sub-sentences are replaced by the sub-sentence which randomly sampled from the documents with the same downstream task label. The third row "train on aspect-k-anonymity" means that the extracted sub-sentences are replaced by the similar aspect sub-sentence through our Aspect $k$-anonymity Substitution process. The fourth and fifth rows "train on original, test on de-identified" settings follows the fidelity setting, so we put them into the fidelity discussion. As we can see, the performance of the third row is close to the first row, which means that our method can preserve most of the utility of the original documents.

Table~\ref{tab:document-fidelity-K-Means-k-8} shows the fidelity of the de-identified summaries. The first and second rows "train on original, test on de-identified" settings shows the performance of cluster agreement. Together with the fourth and fifth rows of Table~\ref{tab:downstream-task-utility}, we can conclude that our method can achieve higher cluster agreement than the random-substitute baseline, which means our method can preserve more statistic properties of the original document distribution. The third and fourth rows follows utility setting, it represents the utility on the down stream task which is a clustering task. In this scenario, our method still outperform random-substitute baseline, which means it can preserve more utility of the original documents on the downstream clustering task.

Table~\ref{tab:ARI-AMI-K-Means-k-8} shows the cluster agreement between "train on original" and "train on de-identified" settings. Although the agreement is quite low, our method still significantly better than the random-substitute baseline.

\subsection{Experiment on Re-identifiability}

In this experiment, we investigate the performance of our framework to de-identify the generate summaries. Inspire from~\citet{morris2022unsupervised}, we train a re-identification model, apply it to the generated de-identified summaries and calculate its re-identify accuracy.

\subsubsection{Setup}

To train the re-identification model, instead of using BERT-style word-level masking, we use a sub-sentence level random sampling method to construct the training data. In order to be consistent to the summarization setting, we randomly sample 10\% of sub-sentences from the original sensitive document and concatenated them as a summary, and train the model to perform classification of 1789 classes (number of persons) with AdamW optimizer and learning rate = $2 \times 10^{-5}$. 

We use the same SBERT model \textit{sentence-transformers/paraphrase-multilingual-MiniLM-L12-v2} as foundation model, and fine-tuning it on the random sampled summary until testing accuracy above 98\%. The testing data are also the random sampled summary.

After fine-tuning, we use the model as the re-identification model and perform prediction on the de-identified summary, calculate its top-1, top5, top-10 and top-100 accuracy.

\subsubsection{Results}
Table~\ref{tab:re-identify-accuracy} shows that the re-identify accuracy of our Aspect $k$-anonymity Substitution method is as good as the random substitute method in all the tested top-n accuracy, which means that it is nearly impossible to re-identify a person from the summary generated by our method.

\section{Conclusion and Future Works}

We have proposed an aspect-based de-identification summarization framework, \textsc{AspirinSum}, to deal with the problem of de-identify indirect identifier without explicit define a category list. Instead, leveraging the fine grain extraction ability of aspect-based summarization framework, we can extract sub-sentences related to the \textit{Personal Sensitive Aspect} (PSA), by giving the domain expert's personal comments. Base on the fact that the domain expert's personal comments usually point out the most salient part of a person, it can be utilized as a source of PSA. With our expert aspect alignment model, XAlign, which utilize cross attention and contrastive learning to mix and align the information from expert's comments and personal sensitive documents, we can learn how expert attending to the personal sensitive document. Our goal is to create a de-identification framework that only utilize existing expert's reference notes data without further human annotations, to make the publishing of sensitive text data easier than before.

Currently, we have explorer our method through many different aspects, including the precision of the aspect extraction, utility and fidelity of the de-identified documents and the risk of being re-identified. However, there are still several directions that we haven't explorer, for example:

\begin{itemize}
    \item The use of splitting by punctuation as sub-sentence chunking mechanism is relatively simple, one might want to use other chunking mechanism, such as \textit{Elementary Discourse Unit} (EDU)~\cite{mann1988rhetorical}, or token level splitting, to see if the precision will be increase.
    \item In the current framework, we only utilize the dense representation of the text, such as sentence embedding. However, many of the sub-sentence chunk contains keyword patterns, can we leverage the prebuild keyword/pattern list to increase the precision/recall of the extracted sub-sentence?
    \item In the utility and fidelity section, we use prediction performance, such as accuracy, to quantify the utility and fidelity of the de-identified text. Can we develop a single score, like IS~\cite{li2017alice} or FID~\cite{heusel2017gans} score in the image generation field, to quantify the quality of the de-identified text?
    \item In the re-identifiability section, we consider using a model trained on the randomly sampled sub-sentences of the sensitive document as the re-identification model. However, this model may be overfitted on the training documents, make the model only rely on the presence of some keywords not on its meaning. Can we use a more robust attacking model, such as membership inference attack~\cite{shokri2017membership}, to quantify the risk of re-identification?
\end{itemize}

Besides, we haven't explorer the quality of LLM generated summary which augmented by the extracted aspect sub-sentences. We leave these directions in our to-do list to make our final result more comprehensive.


\FloatBarrier

\bibliography{aaai22}

\begin{thebibliography}{129}
\providecommand{\natexlab}[1]{#1}

\bibitem[{Abadi et~al.(2016)Abadi, Chu, Goodfellow, McMahan, Mironov, Talwar, and Zhang}]{abadi2016deep}
Abadi, M.; Chu, A.; Goodfellow, I.; McMahan, H.~B.; Mironov, I.; Talwar, K.; and Zhang, L. 2016.
\newblock Deep learning with differential privacy.
\newblock In \emph{Proceedings of the 2016 ACM SIGSAC conference on computer and communications security}, 308--318.

\bibitem[{Abdulateef et~al.(2020)Abdulateef, Khan, Chen, and Shang}]{abdulateef2020multidocument}
Abdulateef, S.; Khan, N.~A.; Chen, B.; and Shang, X. 2020.
\newblock Multidocument Arabic text summarization based on clustering and Word2Vec to reduce redundancy.
\newblock \emph{Information}, 11(2): 59.

\bibitem[{Ahmed, Aziz, and Mohammed(2020)}]{ahmed2020identification}
Ahmed, T.; Aziz, M. M.~A.; and Mohammed, N. 2020.
\newblock De-identification of electronic health record using neural network.
\newblock \emph{Scientific reports}, 10(1): 18600.

\bibitem[{Al~Aziz et~al.(2021)Al~Aziz, Ahmed, Faequa, Jiang, Yao, and Mohammed}]{al2021differentially}
Al~Aziz, M.~M.; Ahmed, T.; Faequa, T.; Jiang, X.; Yao, Y.; and Mohammed, N. 2021.
\newblock Differentially private medical texts generation using generative neural networks.
\newblock \emph{ACM Transactions on Computing for Healthcare (HEALTH)}, 3(1): 1--27.

\bibitem[{Amplayo, Angelidis, and Lapata(2021)}]{amplayo2021aspect}
Amplayo, R.~K.; Angelidis, S.; and Lapata, M. 2021.
\newblock Aspect-controllable opinion summarization.
\newblock \emph{arXiv preprint arXiv:2109.03171}.

\bibitem[{Anandan and Clifton(2011)}]{anandan2011significance}
Anandan, B.; and Clifton, C. 2011.
\newblock Significance of term relationships on anonymization.
\newblock In \emph{2011 IEEE/WIC/ACM International Conferences on Web Intelligence and Intelligent Agent Technology}, volume~3, 253--256. IEEE.

\bibitem[{Angelidis et~al.(2021)Angelidis, Amplayo, Suhara, Wang, and Lapata}]{angelidis2021extractive}
Angelidis, S.; Amplayo, R.~K.; Suhara, Y.; Wang, X.; and Lapata, M. 2021.
\newblock Extractive opinion summarization in quantized transformer spaces.
\newblock \emph{Transactions of the Association for Computational Linguistics}, 9: 277--293.

\bibitem[{Angelidis and Lapata(2018)}]{angelidis2018summarizing}
Angelidis, S.; and Lapata, M. 2018.
\newblock Summarizing opinions: Aspect extraction meets sentiment prediction and they are both weakly supervised.
\newblock \emph{arXiv preprint arXiv:1808.08858}.

\bibitem[{Anjum, Mohammed, and Jiang(2021)}]{anjum2021identification}
Anjum, M.~M.; Mohammed, N.; and Jiang, X. 2021.
\newblock De-identification of unstructured clinical texts from sequence to sequence perspective.
\newblock In \emph{Proceedings of the 2021 ACM SIGSAC Conference on Computer and Communications Security}, 2438--2440.

\bibitem[{Annas(2003)}]{annas2003hipaa}
Annas, G.~J. 2003.
\newblock HIPAA regulations: a new era of medical-record privacy?
\newblock \emph{New England Journal of Medicine}, 348: 1486.

\bibitem[{Bakkelund(2009)}]{bakkelund2009lcs}
Bakkelund, D. 2009.
\newblock An LCS-based string metric.
\newblock \emph{Olso, Norway: University of Oslo}.

\bibitem[{Bansal et~al.(2024)Bansal, Samanta, Dalmia, Gupta, Vashishth, Ganapathy, Bapna, Jain, and Talukdar}]{bansal2024llm}
Bansal, R.; Samanta, B.; Dalmia, S.; Gupta, N.; Vashishth, S.; Ganapathy, S.; Bapna, A.; Jain, P.; and Talukdar, P. 2024.
\newblock Llm augmented llms: Expanding capabilities through composition.
\newblock \emph{arXiv preprint arXiv:2401.02412}.

\bibitem[{Bhaskar, Fabbri, and Durrett(2023)}]{bhaskar-etal-2023-prompted}
Bhaskar, A.; Fabbri, A.; and Durrett, G. 2023.
\newblock Prompted Opinion Summarization with {GPT}-3.5.
\newblock In Rogers, A.; Boyd-Graber, J.; and Okazaki, N., eds., \emph{Findings of the Association for Computational Linguistics: ACL 2023}, 9282--9300. Toronto, Canada: Association for Computational Linguistics.

\bibitem[{Black(2023)}]{Black2023legalassistants}
Black, N. 2023.
\newblock The Future Is Now: The rise of AI-powered legal assistants.
\newblock \url{https://www.abajournal.com/columns/article/the-future-is-now-the-rise-of-ai-powered-legal-assistants}.
\newblock Accessed: 12/15/2023.

\bibitem[{Bougouin, Boudin, and Daille(2013)}]{bougouin2013topicrank}
Bougouin, A.; Boudin, F.; and Daille, B. 2013.
\newblock Topicrank: Graph-based topic ranking for keyphrase extraction.
\newblock In \emph{International joint conference on natural language processing (IJCNLP)}, 543--551.

\bibitem[{Braff(2023)}]{Braff2023ChatGPT}
Braff, D. 2023.
\newblock Some attorneys are using ChatGPT to help them practice more efficiently.
\newblock \url{https://www.americanbar.org/groups/journal/articles/2023/some-attorneys-are-using-chatgpt-help-them-practice-more-efficiently/}.
\newblock Accessed: 12/15/2023.

\bibitem[{Brody and Elhadad(2010)}]{brody2010unsupervised}
Brody, S.; and Elhadad, N. 2010.
\newblock An unsupervised aspect-sentiment model for online reviews.
\newblock In \emph{Human language technologies: The 2010 annual conference of the North American chapter of the association for computational linguistics}, 804--812.

\bibitem[{Brown et~al.(2022)Brown, Lee, Mireshghallah, Shokri, and Tram{\`e}r}]{brown2022does}
Brown, H.; Lee, K.; Mireshghallah, F.; Shokri, R.; and Tram{\`e}r, F. 2022.
\newblock What does it mean for a language model to preserve privacy?
\newblock In \emph{Proceedings of the 2022 ACM Conference on Fairness, Accountability, and Transparency}, 2280--2292.

\bibitem[{Brown et~al.(2020)Brown, Mann, Ryder, Subbiah, Kaplan, Dhariwal, Neelakantan, Shyam, Sastry, Askell et~al.}]{brown2020language}
Brown, T.; Mann, B.; Ryder, N.; Subbiah, M.; Kaplan, J.~D.; Dhariwal, P.; Neelakantan, A.; Shyam, P.; Sastry, G.; Askell, A.; et~al. 2020.
\newblock Language models are few-shot learners.
\newblock \emph{Advances in neural information processing systems}, 33: 1877--1901.

\bibitem[{Carlini et~al.(2019)Carlini, Liu, Erlingsson, Kos, and Song}]{236216}
Carlini, N.; Liu, C.; Erlingsson, {\'U}.; Kos, J.; and Song, D. 2019.
\newblock The Secret Sharer: Evaluating and Testing Unintended Memorization in Neural Networks.
\newblock In \emph{28th USENIX Security Symposium (USENIX Security 19)}, 267--284. Santa Clara, CA: USENIX Association.
\newblock ISBN 978-1-939133-06-9.

\bibitem[{Carlini et~al.(2021)Carlini, Tram{\`e}r, Wallace, Jagielski, Herbert-Voss, Lee, Roberts, Brown, Song, Erlingsson, Oprea, and Raffel}]{274574}
Carlini, N.; Tram{\`e}r, F.; Wallace, E.; Jagielski, M.; Herbert-Voss, A.; Lee, K.; Roberts, A.; Brown, T.; Song, D.; Erlingsson, {\'U}.; Oprea, A.; and Raffel, C. 2021.
\newblock Extracting Training Data from Large Language Models.
\newblock In \emph{30th USENIX Security Symposium (USENIX Security 21)}, 2633--2650. USENIX Association.
\newblock ISBN 978-1-939133-24-3.

\bibitem[{Chakaravarthy et~al.(2008)Chakaravarthy, Gupta, Roy, and Mohania}]{chakaravarthy2008efficient}
Chakaravarthy, V.~T.; Gupta, H.; Roy, P.; and Mohania, M.~K. 2008.
\newblock Efficient techniques for document sanitization.
\newblock In \emph{Proceedings of the 17th ACM conference on Information and knowledge management}, 843--852.

\bibitem[{Chen, Orekondy, and Fritz(2020)}]{chen2020gs}
Chen, D.; Orekondy, T.; and Fritz, M. 2020.
\newblock Gs-wgan: A gradient-sanitized approach for learning differentially private generators.
\newblock \emph{Advances in Neural Information Processing Systems}, 33: 12673--12684.

\bibitem[{Chen et~al.(2020)Chen, Kornblith, Norouzi, and Hinton}]{chen2020simple}
Chen, T.; Kornblith, S.; Norouzi, M.; and Hinton, G. 2020.
\newblock A simple framework for contrastive learning of visual representations.
\newblock In \emph{International conference on machine learning}, 1597--1607. PMLR.

\bibitem[{Chen, Mukherjee, and Liu(2014)}]{chen2014aspect}
Chen, Z.; Mukherjee, A.; and Liu, B. 2014.
\newblock Aspect extraction with automated prior knowledge learning.
\newblock In \emph{Proceedings of the 52nd Annual Meeting of the Association for Computational Linguistics (Volume 1: Long Papers)}, 347--358.

\bibitem[{Chengzhang and Dan(2018)}]{chengzhang2018chinese}
Chengzhang, X.; and Dan, L. 2018.
\newblock Chinese text summarization algorithm based on word2vec.
\newblock In \emph{Journal of Physics: Conference Series}, volume 976, 012006. IOP Publishing.

\bibitem[{Chowdhury, Zhao, and Chaturvedi(2022)}]{chowdhury2022unsupervised}
Chowdhury, S. B.~R.; Zhao, C.; and Chaturvedi, S. 2022.
\newblock Unsupervised extractive opinion summarization using sparse coding.
\newblock \emph{arXiv preprint arXiv:2203.07921}.

\bibitem[{Dalianis(2019)}]{dalianis2019pseudonymisation}
Dalianis, H. 2019.
\newblock Pseudonymisation of Swedish electronic patient records using a rule-based approach.
\newblock In \emph{Proceedings of the Workshop on NLP and Pseudonymisation}, volume 166, 16--23.

\bibitem[{Das and Khetan(2023)}]{das2023deft}
Das, D.; and Khetan, V. 2023.
\newblock DEFT: Data Efficient Fine-Tuning for Large Language Models via Unsupervised Core-Set Selection.
\newblock \emph{arXiv preprint arXiv:2310.16776}.

\bibitem[{Dernoncourt et~al.(2017)Dernoncourt, Lee, Uzuner, and Szolovits}]{dernoncourt2017identification}
Dernoncourt, F.; Lee, J.~Y.; Uzuner, O.; and Szolovits, P. 2017.
\newblock De-identification of patient notes with recurrent neural networks.
\newblock \emph{Journal of the American Medical Informatics Association}, 24(3): 596--606.

\bibitem[{Douglass et~al.(2005)Douglass, Cliffford, Reisner, Long, Moody, and Mark}]{douglass2005identification}
Douglass, M.; Cliffford, G.; Reisner, A.; Long, W.; Moody, G.; and Mark, R. 2005.
\newblock De-identification algorithm for free-text nursing notes.
\newblock In \emph{Computers in Cardiology, 2005}, 331--334. IEEE.

\bibitem[{Douglass et~al.(2004)Douglass, Clifford, Reisner, Moody, and Mark}]{douglass2004computer}
Douglass, M.; Clifford, G.~D.; Reisner, A.; Moody, G.~B.; and Mark, R.~G. 2004.
\newblock Computer-assisted de-identification of free text in the MIMIC II database.
\newblock In \emph{Computers in Cardiology, 2004}, 341--344. IEEE.

\bibitem[{Dwork(2006)}]{dwork2006differential}
Dwork, C. 2006.
\newblock Differential privacy.
\newblock In \emph{International colloquium on automata, languages, and programming}, 1--12. Springer.

\bibitem[{Erkan and Radev(2004)}]{erkan2004lexrank}
Erkan, G.; and Radev, D.~R. 2004.
\newblock Lexrank: Graph-based lexical centrality as salience in text summarization.
\newblock \emph{Journal of artificial intelligence research}, 22: 457--479.

\bibitem[{Estrada(2023)}]{Estrada2023}
Estrada, S. 2023.
\newblock A startup CFO used ChatGPT to build an FP\&A tool—here’s how it went.
\newblock \url{https://fortune.com/2023/03/01/startup-cfo-chatgpt-finance-tool/}.
\newblock Accessed: 09/11/2023.

\bibitem[{{European Parliament} and {Council of the European Union}(2016)}]{http://data.europa.eu/eli/reg/2016/679/oj}
{European Parliament}; and {Council of the European Union}. 2016.
\newblock Regulation (EU) 2016/679 of the European Parliament and of the Council of 27 April 2016 on the protection of natural persons with regard to the processing of personal data and on the free movement of such data, and repealing Directive 95/46/EC (General Data Protection Regulation) (Text with EEA relevance).
\newblock \url{http://data.europa.eu/eli/reg/2016/679/oj}.

\bibitem[{Fabbri et~al.(2021)Fabbri, Wu, Iyer, and Diab}]{fabbri2021multi}
Fabbri, A.~R.; Wu, X.; Iyer, S.; and Diab, M. 2021.
\newblock Multi-perspective abstractive answer summarization.
\newblock \emph{arXiv preprint arXiv:2104.08536}.

\bibitem[{Falke et~al.(2019)Falke, Ribeiro, Utama, Dagan, and Gurevych}]{falke2019ranking}
Falke, T.; Ribeiro, L.~F.; Utama, P.~A.; Dagan, I.; and Gurevych, I. 2019.
\newblock Ranking generated summaries by correctness: An interesting but challenging application for natural language inference.
\newblock In \emph{Proceedings of the 57th annual meeting of the association for computational linguistics}, 2214--2220.

\bibitem[{Ferreira(2023)}]{Ferreira2023}
Ferreira, P. 2023.
\newblock Can ChatGPT Improve Technical Analysis and Trading Techniques?
\newblock \url{https://www.financemagnates.com/trending/can-chatgpt-improve-technical-analysis-and-trading-techniques/}.
\newblock Accessed: 09/11/2023.

\bibitem[{Florescu and Caragea(2017)}]{florescu2017positionrank}
Florescu, C.; and Caragea, C. 2017.
\newblock Positionrank: An unsupervised approach to keyphrase extraction from scholarly documents.
\newblock In \emph{Proceedings of the 55th annual meeting of the association for computational linguistics (volume 1: long papers)}, 1105--1115.

\bibitem[{Fox(2023)}]{Fox2023}
Fox, A. 2023.
\newblock ChatGPT scored 72
\newblock \url{https://www.healthcareitnews.com/news/chatgpt-scored-72-clinical-decision-accuracy-mgb-study-shows}.
\newblock Accessed: 09/11/2023.

\bibitem[{Ganesan, Zhai, and Han(2010)}]{ganesan2010opinosis}
Ganesan, K.; Zhai, C.; and Han, J. 2010.
\newblock Opinosis: A graph based approach to abstractive summarization of highly redundant opinions.
\newblock In \emph{Proceedings of the 23rd international conference on computational linguistics (Coling 2010)}, 340--348.

\bibitem[{Gao, Yao, and Chen(2021)}]{gao2021simcse}
Gao, T.; Yao, X.; and Chen, D. 2021.
\newblock Simcse: Simple contrastive learning of sentence embeddings.
\newblock \emph{arXiv preprint arXiv:2104.08821}.

\bibitem[{Garc{\'\i}a-Pablos, Cuadros, and Rigau(2018)}]{garcia2018w2vlda}
Garc{\'\i}a-Pablos, A.; Cuadros, M.; and Rigau, G. 2018.
\newblock W2VLDA: almost unsupervised system for aspect based sentiment analysis.
\newblock \emph{Expert Systems with Applications}, 91: 127--137.

\bibitem[{Garfinkel et~al.(2015)}]{garfinkel2015identification}
Garfinkel, S.; et~al. 2015.
\newblock \emph{De-identification of Personal Information:.}
\newblock US Department of Commerce, National Institute of Standards and Technology.

\bibitem[{Genest and Lapalme(2012)}]{genest2012fully}
Genest, P.-E.; and Lapalme, G. 2012.
\newblock Fully abstractive approach to guided summarization.
\newblock In \emph{Proceedings of the 50th Annual Meeting of the Association for Computational Linguistics (Volume 2: Short Papers)}, 354--358.

\bibitem[{Germain(2023)}]{Germain2023}
Germain, T. 2023.
\newblock A Mental Health App Tested ChatGPT on Its Users. The Founder Said Backlash Was Just a Misunderstanding.
\newblock \url{https://gizmodo.com/mental-health-therapy-app-ai-koko-chatgpt-rob-morris-1849965534/}.
\newblock Accessed: 09/11/2023.

\bibitem[{Giarelis, Mastrokostas, and Karacapilidis(2023)}]{giarelis2023abstractive}
Giarelis, N.; Mastrokostas, C.; and Karacapilidis, N. 2023.
\newblock Abstractive vs. Extractive Summarization: An Experimental Review.
\newblock \emph{Applied Sciences}, 13(13): 7620.

\bibitem[{Gong and Liu(2001)}]{gong2001generic}
Gong, Y.; and Liu, X. 2001.
\newblock Generic text summarization using relevance measure and latent semantic analysis.
\newblock In \emph{Proceedings of the 24th annual international ACM SIGIR conference on Research and development in information retrieval}, 19--25.

\bibitem[{Haider et~al.(2020)Haider, Hossin, Mahi, and Arif}]{haider2020automatic}
Haider, M.~M.; Hossin, M.~A.; Mahi, H.~R.; and Arif, H. 2020.
\newblock Automatic text summarization using gensim word2vec and k-means clustering algorithm.
\newblock In \emph{2020 IEEE Region 10 Symposium (TENSYMP)}, 283--286. IEEE.

\bibitem[{Harder, Adamczewski, and Park(2021)}]{harder2021dp}
Harder, F.; Adamczewski, K.; and Park, M. 2021.
\newblock Dp-merf: Differentially private mean embeddings with randomfeatures for practical privacy-preserving data generation.
\newblock In \emph{International conference on artificial intelligence and statistics}, 1819--1827. PMLR.

\bibitem[{Hartigan and Wong(1979)}]{hartigan1979algorithm}
Hartigan, J.~A.; and Wong, M.~A. 1979.
\newblock Algorithm AS 136: A k-means clustering algorithm.
\newblock \emph{Journal of the royal statistical society. series c (applied statistics)}, 28(1): 100--108.

\bibitem[{He et~al.(2017)He, Lee, Ng, and Dahlmeier}]{he2017unsupervised}
He, R.; Lee, W.~S.; Ng, H.~T.; and Dahlmeier, D. 2017.
\newblock An unsupervised neural attention model for aspect extraction.
\newblock In \emph{Proceedings of the 55th Annual Meeting of the Association for Computational Linguistics (Volume 1: Long Papers)}, 388--397.

\bibitem[{Heusel et~al.(2017)Heusel, Ramsauer, Unterthiner, Nessler, and Hochreiter}]{heusel2017gans}
Heusel, M.; Ramsauer, H.; Unterthiner, T.; Nessler, B.; and Hochreiter, S. 2017.
\newblock Gans trained by a two time-scale update rule converge to a local nash equilibrium.
\newblock \emph{Advances in neural information processing systems}, 30.

\bibitem[{Hsiao et~al.(2022)Hsiao, Tsai, Liou, and Chen}]{hsiao2022generate}
Hsiao, C.-S.; Tsai, J.-A.; Liou, J.-S.; and Chen, Y.-S. 2022.
\newblock Generate Multi-Perspective Summarization with Pairwise Alignment Mechanism.
\newblock In \emph{2022 International Conference on Technologies and Applications of Artificial Intelligence (TAAI)}, 83--88. IEEE.

\bibitem[{Hu et~al.(2023)Hu, Wu, Li, Long, Garrido, Ge, Ding, Forsyth, Li, and Song}]{hu2023sok}
Hu, Y.; Wu, F.; Li, Q.; Long, Y.; Garrido, G.~M.; Ge, C.; Ding, B.; Forsyth, D.; Li, B.; and Song, D. 2023.
\newblock Sok: Privacy-preserving data synthesis.
\newblock \emph{arXiv preprint arXiv:2307.02106}.

\bibitem[{Hu et~al.(2019)Hu, Havrylov, Titov, and Cohen}]{hu2019obfuscation}
Hu, Z.; Havrylov, S.; Titov, I.; and Cohen, S.~B. 2019.
\newblock Obfuscation for privacy-preserving syntactic parsing.
\newblock \emph{arXiv preprint arXiv:1904.09585}.

\bibitem[{Javaid et~al.(2023)Javaid, Haleem, Singh, Khan, and Khan}]{JAVAID2023100115}
Javaid, M.; Haleem, A.; Singh, R.~P.; Khan, S.; and Khan, I.~H. 2023.
\newblock Unlocking the opportunities through ChatGPT Tool towards ameliorating the education system.
\newblock \emph{BenchCouncil Transactions on Benchmarks, Standards and Evaluations}, 3(2): 100115.

\bibitem[{Jiang et~al.(2023)Jiang, Wang, Wei, Li, and Wang}]{jiang2023large}
Jiang, H.; Wang, R.; Wei, Z.; Li, Y.; and Wang, X. 2023.
\newblock Large-Scale and Multi-Perspective Opinion Summarization with Diverse Review Subsets.
\newblock \emph{arXiv preprint arXiv:2310.13340}.

\bibitem[{Jin, Ho, and Srihari(2009)}]{jin2009opinionminer}
Jin, W.; Ho, H.~H.; and Srihari, R.~K. 2009.
\newblock OpinionMiner: a novel machine learning system for web opinion mining and extraction.
\newblock In \emph{Proceedings of the 15th ACM SIGKDD international conference on Knowledge discovery and data mining}, 1195--1204.

\bibitem[{Jung et~al.(2019)Jung, Kang, Mentch, and Hovy}]{jung2019earlier}
Jung, T.; Kang, D.; Mentch, L.; and Hovy, E. 2019.
\newblock Earlier Isn't Always Better: Sub-aspect Analysis on Corpus and System Biases in Summarization.
\newblock \emph{arXiv preprint arXiv:1908.11723}.

\bibitem[{Keeler and Rumelhart(1991)}]{keeler1991self}
Keeler, J.; and Rumelhart, D. 1991.
\newblock A self-organizing integrated segmentation and recognition neural net.
\newblock \emph{Advances in neural information processing systems}, 4.

\bibitem[{Khan et~al.(2018)Khan, Salim, Farman, Khan, Jan, Ahmad, Ahmed, and Paul}]{khan2018abstractive}
Khan, A.; Salim, N.; Farman, H.; Khan, M.; Jan, B.; Ahmad, A.; Ahmed, I.; and Paul, A. 2018.
\newblock Abstractive text summarization based on improved semantic graph approach.
\newblock \emph{International Journal of Parallel Programming}, 46: 992--1016.

\bibitem[{Kimmel(2023)}]{kimmel2023chatgpt}
Kimmel, D. 2023.
\newblock ChatGPT Therapy Is Good, But It Misses What Makes Us Human.
\newblock \url{https://www.columbiapsychiatry.org/news/chatgpt-therapy-is-good-but-it-misses-what-makes-us-human}.
\newblock Accessed: 09/11/2023.

\bibitem[{Lee(2018)}]{lee2018natural}
Lee, S.~H. 2018.
\newblock Natural language generation for electronic health records.
\newblock \emph{NPJ digital medicine}, 1(1): 63.

\bibitem[{Lehman et~al.(2021)Lehman, Jain, Pichotta, Goldberg, and Wallace}]{lehman2021does}
Lehman, E.; Jain, S.; Pichotta, K.; Goldberg, Y.; and Wallace, B.~C. 2021.
\newblock Does BERT pretrained on clinical notes reveal sensitive data?
\newblock \emph{arXiv preprint arXiv:2104.07762}.

\bibitem[{Leonard(2023)}]{Leonard2023}
Leonard, A. 2023.
\newblock ‘Dr. Google’ meets its match: Dr. ChatGPT.
\newblock \url{https://www.latimes.com/science/story/2023-09-08/dr-google-meets-its-match-dr-chatgpt}.
\newblock Accessed: 09/11/2023.

\bibitem[{Levandowsky and Winter(1971)}]{levandowsky1971distance}
Levandowsky, M.; and Winter, D. 1971.
\newblock Distance between sets.
\newblock \emph{Nature}, 234(5323): 34--35.

\bibitem[{Lewis et~al.(2019)Lewis, Liu, Goyal, Ghazvininejad, Mohamed, Levy, Stoyanov, and Zettlemoyer}]{lewis2019bart}
Lewis, M.; Liu, Y.; Goyal, N.; Ghazvininejad, M.; Mohamed, A.; Levy, O.; Stoyanov, V.; and Zettlemoyer, L. 2019.
\newblock Bart: Denoising sequence-to-sequence pre-training for natural language generation, translation, and comprehension.
\newblock \emph{arXiv preprint arXiv:1910.13461}.

\bibitem[{Li et~al.(2017)Li, Liu, Chen, Pu, Chen, Henao, and Carin}]{li2017alice}
Li, C.; Liu, H.; Chen, C.; Pu, Y.; Chen, L.; Henao, R.; and Carin, L. 2017.
\newblock Alice: Towards understanding adversarial learning for joint distribution matching.
\newblock \emph{Advances in neural information processing systems}, 30.

\bibitem[{Li et~al.(2010)Li, Han, Huang, Zhu, Xia, Zhang, and Yu}]{li2010structure}
Li, F.; Han, C.; Huang, M.; Zhu, X.; Xia, Y.; Zhang, S.; and Yu, H. 2010.
\newblock Structure-aware review mining and summarization.
\newblock In \emph{Proceedings of the 23rd international conference on computational linguistics (Coling 2010)}, 653--661.

\bibitem[{Li, Thadani, and Stent(2016)}]{li2016role}
Li, J.~J.; Thadani, K.; and Stent, A. 2016.
\newblock The role of discourse units in near-extractive summarization.
\newblock In \emph{Proceedings of the 17th Annual Meeting of the Special Interest Group on Discourse and Dialogue}, 137--147.

\bibitem[{Lin(2004)}]{lin2004rouge}
Lin, C.-Y. 2004.
\newblock Rouge: A package for automatic evaluation of summaries.
\newblock In \emph{Text summarization branches out}, 74--81.

\bibitem[{Lison et~al.(2021)Lison, Pil{\'a}n, S{\'a}nchez, Batet, and {\O}vrelid}]{lison2021anonymisation}
Lison, P.; Pil{\'a}n, I.; S{\'a}nchez, D.; Batet, M.; and {\O}vrelid, L. 2021.
\newblock Anonymisation models for text data: State of the art, challenges and future directions.
\newblock In \emph{Proceedings of the 59th Annual Meeting of the Association for Computational Linguistics and the 11th International Joint Conference on Natural Language Processing (Volume 1: Long Papers)}, 4188--4203.

\bibitem[{Liu, Joty, and Meng(2015)}]{liu2015fine}
Liu, P.; Joty, S.; and Meng, H. 2015.
\newblock Fine-grained opinion mining with recurrent neural networks and word embeddings.
\newblock In \emph{Proceedings of the 2015 conference on empirical methods in natural language processing}, 1433--1443.

\bibitem[{Liu et~al.(2016)Liu, Liu, Zhang, Kim, and Gao}]{liu2016improving}
Liu, Q.; Liu, B.; Zhang, Y.; Kim, D.~S.; and Gao, Z. 2016.
\newblock Improving opinion aspect extraction using semantic similarity and aspect associations.
\newblock In \emph{Proceedings of the AAAI conference on artificial intelligence}, volume~30.

\bibitem[{Liu et~al.(2023)Liu, Yu, Zhang, Wu, Cao, Dai, Zhao, Liu, Shen, Li et~al.}]{liu2023deid}
Liu, Z.; Yu, X.; Zhang, L.; Wu, Z.; Cao, C.; Dai, H.; Zhao, L.; Liu, W.; Shen, D.; Li, Q.; et~al. 2023.
\newblock Deid-gpt: Zero-shot medical text de-identification by gpt-4.
\newblock \emph{arXiv preprint arXiv:2303.11032}.

\bibitem[{Locatello et~al.(2019)Locatello, Bauer, Lucic, Raetsch, Gelly, Sch{\"o}lkopf, and Bachem}]{locatello2019challenging}
Locatello, F.; Bauer, S.; Lucic, M.; Raetsch, G.; Gelly, S.; Sch{\"o}lkopf, B.; and Bachem, O. 2019.
\newblock Challenging common assumptions in the unsupervised learning of disentangled representations.
\newblock In \emph{international conference on machine learning}, 4114--4124. PMLR.

\bibitem[{Luhn(1958)}]{luhn1958automatic}
Luhn, H.~P. 1958.
\newblock The automatic creation of literature abstracts.
\newblock \emph{IBM Journal of research and development}, 2(2): 159--165.

\bibitem[{Mann and Thompson(1988)}]{mann1988rhetorical}
Mann, W.~C.; and Thompson, S.~A. 1988.
\newblock Rhetorical structure theory: Toward a functional theory of text organization.
\newblock \emph{Text-interdisciplinary Journal for the Study of Discourse}, 8(3): 243--281.

\bibitem[{Marujo et~al.(2015)Marujo, Port{\^e}lo, Ling, de~Matos, Neto, Gershman, Carbonell, Trancoso, and Raj}]{marujo2015privacy}
Marujo, L.; Port{\^e}lo, J.; Ling, W.; de~Matos, D.~M.; Neto, J.~P.; Gershman, A.; Carbonell, J.; Trancoso, I.; and Raj, B. 2015.
\newblock Privacy-preserving multi-document summarization.
\newblock \emph{arXiv preprint arXiv:1508.01420}.

\bibitem[{Melamud and Shivade(2019)}]{melamud2019towards}
Melamud, O.; and Shivade, C. 2019.
\newblock Towards automatic generation of shareable synthetic clinical notes using neural language models.
\newblock \emph{arXiv preprint arXiv:1905.07002}.

\bibitem[{Miao et~al.(2020)Miao, Li, Wang, and Tan}]{miao2020snippext}
Miao, Z.; Li, Y.; Wang, X.; and Tan, W.-C. 2020.
\newblock Snippext: Semi-supervised opinion mining with augmented data.
\newblock In \emph{Proceedings of The Web Conference 2020}, 617--628.

\bibitem[{Mihalcea and Tarau(2004)}]{mihalcea2004textrank}
Mihalcea, R.; and Tarau, P. 2004.
\newblock Textrank: Bringing order into text.
\newblock In \emph{Proceedings of the 2004 conference on empirical methods in natural language processing}, 404--411.

\bibitem[{Mitchell et~al.(2013)Mitchell, Aguilar, Wilson, and Van~Durme}]{mitchell2013open}
Mitchell, M.; Aguilar, J.; Wilson, T.; and Van~Durme, B. 2013.
\newblock Open domain targeted sentiment.
\newblock In \emph{Proceedings of the 2013 conference on empirical methods in natural language processing}, 1643--1654.

\bibitem[{Morris et~al.(2022)Morris, Chiu, Zabih, and Rush}]{morris2022unsupervised}
Morris, J.~X.; Chiu, J.~T.; Zabih, R.; and Rush, A.~M. 2022.
\newblock Unsupervised Text Deidentification.
\newblock \emph{arXiv preprint arXiv:2210.11528}.

\bibitem[{Mukherjee et~al.(2023)Mukherjee, Hou, Lanfredi, and Summers}]{mukherjee2023feasibility}
Mukherjee, P.; Hou, B.; Lanfredi, R.~B.; and Summers, R.~M. 2023.
\newblock Feasibility of using the privacy-preserving large language model Vicuna for labeling radiology reports.
\newblock \emph{Radiology}, 309(1): e231147.

\bibitem[{Nasr et~al.(2023)Nasr, Carlini, Hayase, Jagielski, Cooper, Ippolito, Choquette-Choo, Wallace, Tram{\`e}r, and Lee}]{nasr2023scalable}
Nasr, M.; Carlini, N.; Hayase, J.; Jagielski, M.; Cooper, A.~F.; Ippolito, D.; Choquette-Choo, C.~A.; Wallace, E.; Tram{\`e}r, F.; and Lee, K. 2023.
\newblock Scalable Extraction of Training Data from (Production) Language Models.
\newblock \emph{arXiv preprint arXiv:2311.17035}.

\bibitem[{Neamatullah et~al.(2008)Neamatullah, Douglass, Lehman, Reisner, Villarroel, Long, Szolovits, Moody, Mark, and Clifford}]{neamatullah2008automated}
Neamatullah, I.; Douglass, M.~M.; Lehman, L.-W.~H.; Reisner, A.; Villarroel, M.; Long, W.~J.; Szolovits, P.; Moody, G.~B.; Mark, R.~G.; and Clifford, G.~D. 2008.
\newblock Automated de-identification of free-text medical records.
\newblock \emph{BMC medical informatics and decision making}, 8(1): 1--17.

\bibitem[{Nisker(2006)}]{nisker2006pipeda}
Nisker, J. 2006.
\newblock Pipeda: A constitutional analysis.
\newblock \emph{Can. B. Rev.}, 85: 317.

\bibitem[{OpenAI(2023)}]{openai2023gpt4}
OpenAI. 2023.
\newblock GPT-4 Technical Report.
\newblock arXiv:2303.08774.

\bibitem[{Patsakis and Lykousas(2023)}]{patsakis2023man}
Patsakis, C.; and Lykousas, N. 2023.
\newblock Man vs the machine: The Struggle for Effective Text Anonymisation in the Age of Large Language Models.
\newblock \emph{arXiv preprint arXiv:2303.12429}.

\bibitem[{Qiu et~al.(2011)Qiu, Liu, Bu, and Chen}]{qiu2011opinion}
Qiu, G.; Liu, B.; Bu, J.; and Chen, C. 2011.
\newblock Opinion word expansion and target extraction through double propagation.
\newblock \emph{Computational linguistics}, 37(1): 9--27.

\bibitem[{Radford et~al.(2019)Radford, Wu, Child, Luan, Amodei, Sutskever et~al.}]{radford2019language}
Radford, A.; Wu, J.; Child, R.; Luan, D.; Amodei, D.; Sutskever, I.; et~al. 2019.
\newblock Language models are unsupervised multitask learners.
\newblock \emph{OpenAI blog}, 1(8): 9.

\bibitem[{Raffel et~al.(2020)Raffel, Shazeer, Roberts, Lee, Narang, Matena, Zhou, Li, and Liu}]{raffel2020exploring}
Raffel, C.; Shazeer, N.; Roberts, A.; Lee, K.; Narang, S.; Matena, M.; Zhou, Y.; Li, W.; and Liu, P.~J. 2020.
\newblock Exploring the limits of transfer learning with a unified text-to-text transformer.
\newblock \emph{The Journal of Machine Learning Research}, 21(1): 5485--5551.

\bibitem[{Reddy and Knight(2016)}]{reddy2016obfuscating}
Reddy, S.; and Knight, K. 2016.
\newblock Obfuscating gender in social media writing.
\newblock In \emph{Proceedings of the First Workshop on NLP and Computational Social Science}, 17--26.

\bibitem[{Reimers and Gurevych(2019)}]{reimers-2019-sentence-bert}
Reimers, N.; and Gurevych, I. 2019.
\newblock Sentence-BERT: Sentence Embeddings using Siamese BERT-Networks.
\newblock In \emph{Proceedings of the 2019 Conference on Empirical Methods in Natural Language Processing}. Association for Computational Linguistics.

\bibitem[{Reimers and Gurevych(2020)}]{reimers-2020-multilingual-sentence-bert}
Reimers, N.; and Gurevych, I. 2020.
\newblock Making Monolingual Sentence Embeddings Multilingual using Knowledge Distillation.
\newblock In \emph{Proceedings of the 2020 Conference on Empirical Methods in Natural Language Processing}. Association for Computational Linguistics.

\bibitem[{Rekabdar, Mousas, and Gupta(2019)}]{rekabdar2019generative}
Rekabdar, B.; Mousas, C.; and Gupta, B. 2019.
\newblock Generative adversarial network with policy gradient for text summarization.
\newblock In \emph{2019 IEEE 13th international conference on semantic computing (ICSC)}, 204--207. IEEE.

\bibitem[{Rombach et~al.(2022)Rombach, Blattmann, Lorenz, Esser, and Ommer}]{rombach2022high}
Rombach, R.; Blattmann, A.; Lorenz, D.; Esser, P.; and Ommer, B. 2022.
\newblock High-resolution image synthesis with latent diffusion models.
\newblock In \emph{Proceedings of the IEEE/CVF conference on computer vision and pattern recognition}, 10684--10695.

\bibitem[{Samarati(2001)}]{samarati2001protecting}
Samarati, P. 2001.
\newblock Protecting respondents identities in microdata release.
\newblock \emph{IEEE transactions on Knowledge and Data Engineering}, 13(6): 1010--1027.

\bibitem[{S{\'a}nchez and Batet(2016)}]{sanchez2016c}
S{\'a}nchez, D.; and Batet, M. 2016.
\newblock C-sanitized: A privacy model for document redaction and sanitization.
\newblock \emph{Journal of the Association for Information Science and Technology}, 67(1): 148--163.

\bibitem[{Shi et~al.(2018)Shi, Kang, Choo, and Reddy}]{shi2018short}
Shi, T.; Kang, K.; Choo, J.; and Reddy, C.~K. 2018.
\newblock Short-text topic modeling via non-negative matrix factorization enriched with local word-context correlations.
\newblock In \emph{Proceedings of the 2018 World Wide Web Conference}, 1105--1114.

\bibitem[{Shi et~al.(2021)Shi, Li, Wang, and Reddy}]{shi2021simple}
Shi, T.; Li, L.; Wang, P.; and Reddy, C.~K. 2021.
\newblock A simple and effective self-supervised contrastive learning framework for aspect detection.
\newblock In \emph{Proceedings of the AAAI conference on artificial intelligence}, volume~35, 13815--13824.

\bibitem[{Shokri et~al.(2017)Shokri, Stronati, Song, and Shmatikov}]{shokri2017membership}
Shokri, R.; Stronati, M.; Song, C.; and Shmatikov, V. 2017.
\newblock Membership inference attacks against machine learning models.
\newblock In \emph{2017 IEEE symposium on security and privacy (SP)}, 3--18. IEEE.

\bibitem[{Staab et~al.(2023)Staab, Vero, Balunovi{\'c}, and Vechev}]{staab2023beyond}
Staab, R.; Vero, M.; Balunovi{\'c}, M.; and Vechev, M. 2023.
\newblock Beyond memorization: Violating privacy via inference with large language models.
\newblock \emph{arXiv preprint arXiv:2310.07298}.

\bibitem[{Steinberger, Jezek et~al.(2004)}]{steinberger2004using}
Steinberger, J.; Jezek, K.; et~al. 2004.
\newblock Using latent semantic analysis in text summarization and summary evaluation.
\newblock \emph{Proc. ISIM}, 4(93-100): 8.

\bibitem[{Stubbs, Kotfila, and Uzuner(2015)}]{stubbs2015automated}
Stubbs, A.; Kotfila, C.; and Uzuner, {\"O}. 2015.
\newblock Automated systems for the de-identification of longitudinal clinical narratives: Overview of 2014 i2b2/UTHealth shared task Track 1.
\newblock \emph{Journal of biomedical informatics}, 58: S11--S19.

\bibitem[{Suhara et~al.(2020)Suhara, Wang, Angelidis, and Tan}]{suhara2020opiniondigest}
Suhara, Y.; Wang, X.; Angelidis, S.; and Tan, W.-C. 2020.
\newblock OpinionDigest: A simple framework for opinion summarization.
\newblock \emph{arXiv preprint arXiv:2005.01901}.

\bibitem[{Sweeney(2002)}]{sweeney2002k}
Sweeney, L. 2002.
\newblock k-anonymity: A model for protecting privacy.
\newblock \emph{International journal of uncertainty, fuzziness and knowledge-based systems}, 10(05): 557--570.

\bibitem[{Taver(2023)}]{taver2023chatgpt}
Taver, M. 2023.
\newblock ChatGPT is Coming to Finance, So Let’s Talk About the Risks and Rewards.
\newblock \url{https://www.unite.ai/chatgpt-is-coming-to-finance-so-lets-talk-about-the-risks-and-rewards/}.
\newblock Accessed: 09/11/2023.

\bibitem[{Torkzadehmahani, Kairouz, and Paten(2019)}]{torkzadehmahani2019dp}
Torkzadehmahani, R.; Kairouz, P.; and Paten, B. 2019.
\newblock Dp-cgan: Differentially private synthetic data and label generation.
\newblock In \emph{Proceedings of the IEEE/CVF Conference on Computer Vision and Pattern Recognition Workshops}, 0--0.

\bibitem[{Van Den~Oord, Vinyals et~al.(2017)}]{van2017neural}
Van Den~Oord, A.; Vinyals, O.; et~al. 2017.
\newblock Neural discrete representation learning.
\newblock \emph{Advances in neural information processing systems}, 30.

\bibitem[{Vaswani et~al.(2017)Vaswani, Shazeer, Parmar, Uszkoreit, Jones, Gomez, Kaiser, and Polosukhin}]{vaswani2017attention}
Vaswani, A.; Shazeer, N.; Parmar, N.; Uszkoreit, J.; Jones, L.; Gomez, A.~N.; Kaiser, {\L}.; and Polosukhin, I. 2017.
\newblock Attention is all you need.
\newblock \emph{Advances in neural information processing systems}, 30.

\bibitem[{Veli{\v{c}}kovi{\'c} et~al.(2017)Veli{\v{c}}kovi{\'c}, Cucurull, Casanova, Romero, Lio, and Bengio}]{velivckovic2017graph}
Veli{\v{c}}kovi{\'c}, P.; Cucurull, G.; Casanova, A.; Romero, A.; Lio, P.; and Bengio, Y. 2017.
\newblock Graph attention networks.
\newblock \emph{arXiv preprint arXiv:1710.10903}.

\bibitem[{Wang et~al.(2015)Wang, Liu, Cao, Zhao, and De~Melo}]{wang2015sentiment}
Wang, L.; Liu, K.; Cao, Z.; Zhao, J.; and De~Melo, G. 2015.
\newblock Sentiment-aspect extraction based on restricted boltzmann machines.
\newblock In \emph{Proceedings of the 53rd Annual Meeting of the Association for Computational Linguistics and the 7th International Joint Conference on Natural Language Processing (Volume 1: Long Papers)}, 616--625.

\bibitem[{Wang et~al.(2016)Wang, Pan, Dahlmeier, and Xiao}]{wang2016recursive}
Wang, W.; Pan, S.~J.; Dahlmeier, D.; and Xiao, X. 2016.
\newblock Recursive neural conditional random fields for aspect-based sentiment analysis.
\newblock \emph{arXiv preprint arXiv:1603.06679}.

\bibitem[{Xiao et~al.(2022)Xiao, Miculicich, Liu, He, and Carenini}]{xiao2022attend}
Xiao, W.; Miculicich, L.; Liu, Y.; He, P.; and Carenini, G. 2022.
\newblock Attend to the Right Context: A Plug-and-Play Module for Content-Controllable Summarization.
\newblock \emph{arXiv preprint arXiv:2212.10819}.

\bibitem[{Yang and Cardie(2012)}]{yang2012extracting}
Yang, B.; and Cardie, C. 2012.
\newblock Extracting opinion expressions with semi-markov conditional random fields.
\newblock In \emph{Proceedings of the 2012 Joint Conference on Empirical Methods in Natural Language Processing and Computational Natural Language Learning}, 1335--1345.

\bibitem[{Yang and Garibaldi(2015)}]{yang2015automatic}
Yang, H.; and Garibaldi, J.~M. 2015.
\newblock Automatic detection of protected health information from clinic narratives.
\newblock \emph{Journal of biomedical informatics}, 58: S30--S38.

\bibitem[{Yang et~al.(2020)Yang, Li, Shen, Wu, Zhao, and Chen}]{yang2020hierarchical}
Yang, M.; Li, C.; Shen, Y.; Wu, Q.; Zhao, Z.; and Chen, X. 2020.
\newblock Hierarchical human-like deep neural networks for abstractive text summarization.
\newblock \emph{IEEE Transactions on Neural Networks and Learning Systems}, 32(6): 2744--2757.

\bibitem[{Yang et~al.(2016)Yang, Yang, Dyer, He, Smola, and Hovy}]{yang2016hierarchical}
Yang, Z.; Yang, D.; Dyer, C.; He, X.; Smola, A.; and Hovy, E. 2016.
\newblock Hierarchical attention networks for document classification.
\newblock In \emph{Proceedings of the 2016 conference of the North American chapter of the association for computational linguistics: human language technologies}, 1480--1489.

\bibitem[{Yogarajan, Pfahringer, and Mayo(2020)}]{yogarajan2020review}
Yogarajan, V.; Pfahringer, B.; and Mayo, M. 2020.
\newblock A review of automatic end-to-end de-identification: Is high accuracy the only metric?
\newblock \emph{Applied Artificial Intelligence}, 34(3): 251--269.

\bibitem[{Yogatama, Liu, and Smith(2015)}]{yogatama2015extractive}
Yogatama, D.; Liu, F.; and Smith, N.~A. 2015.
\newblock Extractive summarization by maximizing semantic volume.
\newblock In \emph{Proceedings of the 2015 Conference on Empirical Methods in Natural Language Processing}, 1961--1966.

\bibitem[{Zhang et~al.(2020)Zhang, Zhao, Saleh, and Liu}]{zhang2020pegasus}
Zhang, J.; Zhao, Y.; Saleh, M.; and Liu, P. 2020.
\newblock Pegasus: Pre-training with extracted gap-sentences for abstractive summarization.
\newblock In \emph{International Conference on Machine Learning}, 11328--11339. PMLR.

\bibitem[{Zhang et~al.(2023)Zhang, Zhou, Huang, He, Yu, and Liu}]{zhang2023asu}
Zhang, M.; Zhou, G.; Huang, N.; He, P.; Yu, W.; and Liu, W. 2023.
\newblock AsU-OSum: Aspect-augmented unsupervised opinion summarization.
\newblock \emph{Information Processing \& Management}, 60(1): 103138.

\bibitem[{Zhao and Chaturvedi(2020)}]{zhao2020weakly}
Zhao, C.; and Chaturvedi, S. 2020.
\newblock Weakly-supervised opinion summarization by leveraging external information.
\newblock In \emph{Proceedings of the AAAI Conference on Artificial Intelligence}, volume~34, 9644--9651.

\bibitem[{Zhao, Gui, and He(2023)}]{zhao2023cone}
Zhao, R.; Gui, L.; and He, Y. 2023.
\newblock Cone: Unsupervised Contrastive Opinion Extraction.
\newblock \emph{arXiv preprint arXiv:2305.04599}.

\bibitem[{Zhao et~al.(2010)Zhao, Jiang, Yan, and Li}]{zhao2010jointly}
Zhao, X.; Jiang, J.; Yan, H.; and Li, X. 2010.
\newblock Jointly modeling aspects and opinions with a MaxEnt-LDA hybrid.
\newblock ACL.

\end{thebibliography}




\clearpage

\appendix

\renewcommand{\thesection}{\Alph{section}.}
\renewcommand{\thesubsection}{\Alph{section}.\arabic{subsection}}

\section{Appendix}

\subsection{Example of Extracted Sub-sentences}
\label{app:example_extracted_sub_sentences}

\Cref{tab:sample_doc_id_1XX_230101X8,tab:sample_doc_id_1XX_230106X0} are examples of the extracted sub-sentences with XAlign+ARCSS(threshold=0.5, iter=1) setting. The corresponding documents are sampled from the testing set, together with its expert comments, human labels and predicted labels.




\vspace{10pt} 

\noindent\begin{minipage}{\textwidth}

    \centering
    \begin{tabular}{|m{4cm}|c|c|m{8cm}|}
        \hline
        \textbf{Expert Comments} & \textbf{Label} & \textbf{Predict} & \textbf{Sub-sentence} \\
        \hline
        \multirow{3}{=}{\parbox{4cm}{電機系 \\
        維也納國際青少年絃樂團比賽運動佳參加非主流機器人賽 \\
        機器人 \\
        小提琴 \\
        自學 \\
        世界機關王大賽 \\
        + 課業成績於該校頂尖小提琴方面成就卓越 \\
        建議改以音樂相關管道申請 \\
        }} & 1 & 1 & 2018年世界機器人機關王競賽世界冠軍語文檢定多益TOEIC 860 金證全民英檢中級能力測驗個人簡歷德文檢定A1 \\ 
        \cline{2-4}
        & 1 & 1 & C++與Arduino實作營師大附中電算社教育部化學人才培訓桃園市數理資優培訓北臺灣聯招小提琴第2名歐洲巡迴演出參加維也納國際青少年音樂節 \\
        \cline{2-4}
        & 0 & 1 & 父親從事手機創新與研發工作 \\
        \cline{2-4}
        & 0 & 1 & 母親曾任銀行理財專員 \\
        \cline{2-4}
        & 0 & 1 & 參與各項音樂比賽屢獲佳績 \\
        \cline{2-4}
        & 1 & 0 & 多次市賽亞軍、巴雀盃音樂大賽亞軍、全北臺灣小提琴第二名考上第一志願師大附中 \\
        \cline{2-4}
        & 1 & 1 & 更於維也納國際弦樂團比賽獲得世界亞軍、站上「音樂至高殿堂」維也納金色大廳及維也納國家音樂廳演奏、歐洲巡迴演出多場音樂會 \\
        \cline{2-4}
        & 0 & 1 & 我發現我對機器人與人工智慧很感興趣 \\
        \cline{2-4}
        & 1 & 1 & 並於2018、2019年參加世界機關王大賽 \\
        \cline{2-4}
        & 0 & 1 & 而高中音樂班少了物理、化學課 \\
        \cline{2-4}
        & 0 & 1 & 音樂為輔」開始規劃我未來的求學生涯 \\
        \hline
    \end{tabular}
    \captionof{table}{Random sample from testing set. doc\_id = 1XX-230101X8, grade:F}
    \label{tab:sample_doc_id_1XX_230101X8}

\end{minipage}


\begin{table*}[h]
    \centering
    \renewcommand{\arraystretch}{4.5} 
    \begin{tabular}{|m{4cm}|c|c|m{8cm}|}
        \hline
        \textbf{Expert Comments} & \textbf{Label} & \textbf{Predict} & \textbf{Sub-sentence} \\
        \hline
        \multirow{3}{=}{\parbox{4cm}{類排42\% \\
        電資院學士班 \\
        ITIA 資訊技術與產業應用國際研討會 \\
        發表論文資策會資安技能金盾獎入圍決賽 \\
        Speakada LLC 工作經驗 \\
        通報9個全國國中小校網使用之Tad插件重大CVE漏洞 \\
        參與過資安競賽程式開發多種類型之競賽皆有不錯的成績累積許多前端專案跨足大型專案開發 \\
        製作模擬聯合國會議系統 \\
        台南一中校內美廣點餐系統系統規劃者[Name1]負責前端[Name2]負責後端參與黑客松 \\
        製作職缺媒合平台環境保護志工媒合平台社會關懷部分 \\
        創立了[keyword1]及[keyword2]為各地學子民眾提供資訊服務 \\
        資安專長 \\
        ITIA論文 \\
        點餐系統 \\
        程式開發分散式資料庫增加資料儲存安全性之方法研究 \\
        ITIA \\
        }} & 1 & 0 & 讀書計畫我與朋友、同學創辦[keyword1] 資訊科技分享網站 \\
        \cline{2-4}  
        \cline{2-4}
        & 1 & 0 & 我創辦[keyword2] 筆記共享平臺 \\
        \cline{2-4}
        & 1 & 0 & 投稿至ITIA 國際研討會錄取 \\
        \cline{2-4}
        & 0 & 1 & 自傳我協助學生會製作美食廣場訂餐系統 \\
        \cline{2-4}
        & 0 & 1 & 麻美化文作品集國際研討會發表區域及原住 \\
        \cline{2-4}
        & 1 & 0 & 廣泛將學習到的知識透過創辦的 [keyword1] 進一步向大眾傳播資訊 \\
        \cline{2-4}
        & 1 & 0 & 國中時創辦了[keyword2] \\
        \cline{2-4}
        & 1 & 0 & 3 統整所學並分享我希望能如我先前創辦[keyword2] 時一樣 \\
        \hline
    \end{tabular}
    \caption{Random sample from testing set. doc\_id = 1XX-230106X0, grade:A}
    \label{tab:sample_doc_id_1XX_230106X0}
\end{table*}


\end{CJK}
\end{document}